\documentclass{article}

\usepackage{microtype}
\usepackage{graphicx}
\usepackage{booktabs} \usepackage{commands}
\usepackage{xcolor}
\usepackage{multirow}
\usepackage{array}
\usepackage{diagbox}
\usepackage{subcaption}
\usepackage{cancel}
\usepackage{tikz}
\usepackage[ruled,vlined]{algorithm2e}
\SetArgSty{textnormal}
\SetKwProg{Process}{Process}{}{}
\SetKwProg{Procedure}{Procedure}{}{}
\SetKwFunction{Encode}{Encode}{}{}
\SetKwFunction{Decode}{Decode}{}{}

\makeatletter
\g@addto@macro{\@algocf@init}{\SetKwInOut{Require}{Require}}
\makeatother

\usepackage{hyperref}

\usepackage[accepted]{icml2021}

\usepackage{mathtools}
\newcommand\given{\,\vert\,}

\DeclarePairedDelimiterX{\divergence}[2]{(}{)}{#1\,\delimsize\|\,#2}
\newcommand{\kl}{D_\mathrm{KL}\divergence}

\usepackage[normalem]{ulem}

\newcommand{\symbolspace}{\sS}
\newcommand{\latentsymbolspace}{\sS'}

\icmltitlerunning{Monte Carlo Bits-Back Coding}

\begin{document}

\twocolumn[
\icmltitle{Improving Lossless Compression Rates via Monte Carlo Bits-Back Coding}

\icmlsetsymbol{equal}{*}

\begin{icmlauthorlist}
\icmlauthor{Yangjun Ruan}{equal,to,vector}
\icmlauthor{Karen Ullrich}{equal,vector,fb}
\icmlauthor{Daniel Severo}{equal,to,vector}
\icmlauthor{James Townsend}{ucl}
\icmlauthor{Ashish Khisti}{to}
\icmlauthor{Arnaud Doucet}{oxford}
\icmlauthor{Alireza Makhzani}{to,vector}
\icmlauthor{Chris J. Maddison}{to,vector}
\end{icmlauthorlist}

\icmlaffiliation{to}{University of Toronto}
\icmlaffiliation{fb}{Facebook AI Research}
\icmlaffiliation{vector}{Vector Institute}
\icmlaffiliation{oxford}{University of Oxford}
\icmlaffiliation{ucl}{University College London}

\icmlcorrespondingauthor{Yangjun Ruan, Daniel Severo, Chris Maddison}{yjruan@cs.toronto.edu, d.severo@mail.utoronto.ca, cmaddis@cs.toronto.edu}

\icmlkeywords{Machine Learning, ICML}

\vskip 0.3in
]

\printAffiliationsAndNotice{\icmlEqualContribution}

\begin{abstract}
Latent variable models have been successfully applied in lossless compression with the bits-back coding algorithm. However, bits-back suffers from an increase in the bitrate equal to the KL divergence between the approximate posterior and the true posterior. In this paper, we show how to remove this gap asymptotically by deriving bits-back coding algorithms from tighter variational bounds. The key idea is to exploit extended space representations of Monte Carlo estimators of the marginal likelihood. Naively applied, our schemes would require more initial bits than the standard bits-back coder, but we show how to drastically reduce this additional cost with couplings in the latent space. When parallel architectures can be exploited, our coders can achieve better rates than bits-back with little additional cost. We demonstrate improved lossless compression rates in a variety of settings, especially in out-of-distribution or sequential data compression.
\end{abstract}
\section{Introduction}

Datasets keep getting bigger; the recent CLIP model was trained on 400 million text-image pairs gathered from the internet \citep{clipopenai}. With datasets of this size coming from ever more heterogeneous sources, we need compression algorithms that can store this data efficiently.

In principle, data compression can be improved with a better approximation of the data generating distribution. Luckily, the quality of generative models is rapidly improving \citep{wavenet, salimans2017pixelcnn, razavi2019generating, vahdat2020nvae}. From this panoply of generative models, latent variable models are particularly attractive for compression applications, because they are typically easy to parallelize; speed is a major concern for compression methods. Indeed, some of the most successful learned compressors for large scale natural images are based on deep latent variable models (see \citet{yang2020improving} for lossy, and \citet{townsend2020hilloc} for lossless).

Lossless compression with latent variable models has a complication that must be addressed. Any model-based coder needs to evaluate the model's probability mass function $p(x)$. Latent variable models are specified in terms of a joint probability mass function $p(x, z)$, where $z$ is a latent, unobserved variable. Computing $p(x)$ in these models requires a (typically intractable) summation, and achieving the model's optimal bitrate, $-\log p(x)$, is not always feasible. When compressing large datasets of i.i.d.\ data, it is possible to approximate this optimal bitrate using the bits-back coding algorithm  \cite{hinton1993keeping, townsend2019practical}. Bits-back coding is based on variational inference \cite{jordan1999introduction}, which uses an approximation $q(z \given x)$ to the true posterior $p(z \given x)$. Unfortunately, this adds roughly $\kl{q(z \given x)}{p(z \given x)}$ bits to the bitrate. This seems unimprovable for a fixed $q$, which is a problem, if approximating $p(z \given x)$ is difficult or expensive.

\begin{figure}[t]
    \centering
    \includegraphics{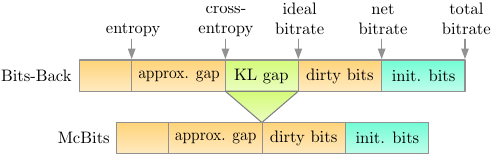}
    \caption{Monte Carlo Bits-Back coders reduce the KL gap to zero.}
    \label{fig:bitrates}
    \vspace{-\baselineskip}
\end{figure}

In this paper, we show how to remove (asymptotically) the $\KL$ gap of bits-back schemes for (just about) any fixed $q$. Our method is based on recent work that derives tighter variational bounds using Monte Carlo estimators of the marginal likelihood \citep[e.g.,][]{burda2015importance}. The idea is that $q$ and $p$ can be lifted into an extended latent space \citep[e.g.,][]{andrieu2010particle} such that the $\KL$ over the extended latent space goes to zero and the overall bitrate approaches $-\log p(x)$. For example, our simplest extended bits-back method, based on importance sampling, introduces $N$ identically distributed particles $z_i$ and a categorical random variable that picks from $z_i$ to approximate $p(z \given x)$. We also define extended bits-back schemes based on more advanced Monte Carlo methods (AIS, \citeauthor{neal2001annealed}, \citeyear{neal2001annealed}; SMC, \citeauthor{doucet2001introduction}, \citeyear{doucet2001introduction}).

Adding $\gO(N)$ latent variables introduces another challenge that we show how to address. Bits-back requires an initial source of bits, and, naively applied, our methods increase the initial bit cost by $\gO(N)$. One of our key contributions is to show that this cost can be reduced to $\gO(\log N)$ for some of our coders using couplings in latent space, a novel technique that may be applicable in other settings to reduce initial bit costs. Most of our coders can be parallelized over the number of particles, which significantly reduces the computation overhead. So, our coders extend bits-back with little additional cost.

We test our methods in various lossless compression settings, including compression of natural images and musical pieces using deep latent variable models. We report between  2\% - 19\% rate savings in our experiments, and we see our most significant improvements when compressing out-of-distribution data or sequential data. We also show that our methods can be used to improve the entropy coding step of learned transform coders with hyperpriors \citep{balle2018variational} for lossy compression. We explore the factors that affect the rate savings in our setting.

\section{Background}

\subsection{Asymmetric Numeral Systems}\label{ANS}
The goal of lossless compression is to find a short binary representation of the outcome of a discrete random variable $x \sim p_d(x)$ in a finite symbol space $x \in \symbolspace$. Achieving the best possible expected length, i.e., the entropy $H(p_d)$ of $p_d$, requires access to $p_d$, and typically a \emph{model} probability mass function (PMF) $p(x)$ is used instead. In this case, the smallest achievable length is the \emph{cross-entropy} of $p$ relative to $p_d$, $H(p_d, p) = - \sum_{x} p_d(x)\log p(x)$\footnote{All logarithms in this paper are base 2.}. See \citet{mackay2003information,cover1991} for more detail.

\emph{Asymmetric numeral systems} (ANS) are model-based coders that achieve near optimal compression rates on sequences of symbols \citep{duda2009asymmetric}. ANS stores data in a stack-like `message' data structure, which we denote $m$, and provides an inverse pair of functions, $\mathrm{encode}_p$ and $\mathrm{decode}_p$, which each process one symbol $x \in \symbolspace$ at a time:
\begin{equation}
\begin{aligned}
    &\mathrm{encode}_p:m, x\mapsto m'\\
    &\mathrm{decode}_p:m'\mapsto (m, x).
\end{aligned}
\end{equation}
Encode pushes $x$ onto $m$, and decode pops $x$ from $m'$. Both functions require access to routines for computing the cumulative distribution function (CDF) and inverse CDF of $p$. If $n$ symbols, drawn i.i.d.\ from $p_d$, are pushed onto $m$ with $p$ , the \emph{bitrate} (bits/symbol) required to store the ANS message approaches $H(p_d, p) + \epsilon$ for some small error $\epsilon$ \citep{duda2009asymmetric, townsend2020tutorial}.
The exact sequence is recovered by popping $n$ symbols off the final $m$ with $p$.

\begin{figure}[t]
    \centering
    \includegraphics{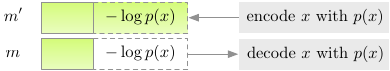}
    \caption{ANS is a last-in-first-out lossless coder. We adopt the visualizations of \citep{kingma2019bit}: the green bars represent the message $m$, a stack that stores symbols $x$.}
    \label{fig:ans}
    \vspace{-.5\baselineskip}
\end{figure}

For our purposes, the ANS message can be thought of as a store of randomness. Given a message $m$ with enough bits, regardless of $m$'s provenance, we can decode from $m$ using any distribution $p$. This will return a random symbol $x$ and remove roughly $-\log p(x)$ bits from $m$. Conversely, we can encode a symbol $x$ onto $m$ with $p$, which will increase $m$'s length by roughly $-\log p(x)$. If a sender produces a message $m$ through some sequence of encode or decode steps using distributions $p_i$, then a receiver, who has access to the $p_i$, can recover the sequence of encoded symbols and the initial message by reversing the order of operations and switching encodes with decodes. When describing algorithms, we often leave out explicit references to $m$, instead writing steps like `$\mathrm{encode} \ x$ with $p(x)$'. See Fig. \ref{fig:ans}.

\subsection{Bits-Back Compression with ANS}

The class of latent variable models is highly flexible, and most operations required to compute with such models can be parallelized, making them an attractive choice for model-based coders. Bits-back coders, in particular Bits-Back with ANS \citep[BB-ANS,][]{townsend2019practical}, specialize in compression using latent variable models.

A latent variable model is specified in terms of a {joint} distribution $p(x, z)$ between $x$ and a latent discrete\footnote{BB-ANS can easily be extended to continuous $z$, with negligible cost, by quantizing; see \citet{townsend2019practical}.} random variable taking value in a symbol space $z \in \latentsymbolspace$. We assume that the joint distribution of latent variable models factorizes as  $p(z)p(x \given z)$ and that the PMFs, CDFs, and inverse CDFs, under $p(z)$ and $p(x \given z)$ are tractable. However, computing the marginal $p(x) = \sum_{z} p(z) p(x \given z)$ is often intractable.  This fact means that we cannot directly encode $x$ onto $m$. A naive strategy would be for the sender to pick some $z \in \latentsymbolspace$, and encode $(x, z)$ using $p$, which would require $-\log p(x, z)$ bits; however, this involves communicating the symbol $z$, which is redundant information.

\begin{figure}[t]
    \centering
    \includegraphics{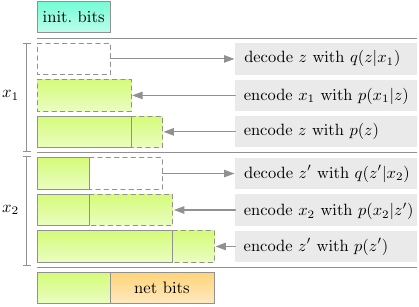}
    \caption{Encoding $(x_1, x_2)$ requires an initial source of bits (light blue), but bits-back reduces its total bit consumption by using intermediate messages as the initial source of bits for encoding $x_2$. The net bits used (light orange) is close to the negative ELBO.}
    \label{fig:bbans}
\end{figure}

BB-ANS gets a better bitrate, by compressing sequences of symbols in a chain and by having the sender \emph{decode} latents $z$ from the intermediate message state, rather than picking $z$; see Fig. \ref{fig:bbans}. Suppose that we have already pushed some symbols onto a message $m$. BB-ANS uses an approximate posterior $q(z \given x)$ such that if $p(x, z) = 0$ then $q(z \given x) = 0$. To encode a symbol $x$ onto $m$, the sender first pops $z$ from $m$ using $q(z \given x)$. Then they push $(x, z)$ onto $m$ using $p(x, z)$. The new message $m'$ has approximately $-\log p(x, z) + \log q(z \given x)$ more bits than $m$. $m'$ is then used in exactly the same way for the next symbol. The per-symbol rate saving over the naive method is $-\log q(z \given x)$ bits. However, for the first symbol, an initial message is needed, causing a one-time overhead.

\subsection{The Bitrate of Bits-Back Coders}

 When encoding a sequence of symbols, we define the \emph{total bitrate} to be the number of bits in the final message per symbol encoded; the \emph{initial bits} to be the number of bits needed to initialize the message; and the \emph{net bitrate} to be the total bitrate minus the initial bits per symbol, which is equal to the expected increase in message length per symbol. As the number of encoded symbols grows, the total bitrate of BB-ANS will converge to the net bitrate.

One subtlety is that the BB-ANS bitrate depends on the distribution of the latent $z$, popped with $q(z\given x)$. In an ideal scenario, where $m$ contains i.i.d.\ uniform Bernoulli distributed bits, $z$ will be an exact sample from $q$. Unfortunately, in practical situations, the exact distribution of $z$ is difficult to characterize. Nevertheless, \citet{townsend2019practical} found the `evidence lower bound' (ELBO)
\begin{equation}\label{eq:elbo}
\begin{aligned}
  &\expect_{z\sim q(z\given x)}\left[-\log p(x, z) + \log q(z \given x)\right]\\
  &= - \log p(x) + \kl{q(z \given x)}{p(z \given x)},
\end{aligned}
\end{equation}
which assumes $z \sim q(z\given x)$, to be an accurate predictor of BB-ANS's empirical compression rate; the effect of inaccurate samples (which they refer to as `dirty bits') is typically less than $1\%$. So, in this paper we mostly elide the dirty bits issue, regarding (\ref{eq:elbo}) to be the net bitrate of BB-ANS, and hereafter we refer to vanilla BB-ANS as BB-ELBO. Taking the expectation under $p_d$, BB-ELBO achieves a net bitrate of approximately $H(p_d, p) +  \mathbb{E}_{x\sim p_d}[\kl{q(z \given x)}{p(z \given x)}]$.

\begin{algorithm*}[t]
\caption{Extended Latent Space Representation of Importance Sampling}
\label{alg:is_extend_rep}

\begin{minipage}[t]{0.46\textwidth}
\Process{$Q(\gZ \given x)$}{
    sample $\{z_i\}_{i=1}^N \sim \prod_{i=1}^N q(z_i \given x)$ \\
    compute $\tilde{w}_i \propto \frac{p(x, z_i)}{q(z_i \given x)}$\\
    sample $j \sim \cat\left(\tilde{w}_j\right)$ \\
    \KwRet $\{z_i\}_{i=1}^N, j$
}
\end{minipage}
\begin{minipage}[t]{0.46\textwidth}
\Process{$P(x, \gZ)$}{
    sample $j \sim \cat(1/N)$ \\
    sample $z_j \sim p(z_j)$ \\
    sample $x \sim p(x \given z_j)$ \\
    sample $\{z_i\}_{i \neq j} \sim \prod_{i \neq j}q(z_i \given x)$ \\
    \KwRet $x, \{z_i\}_{i=1}^N, j$
}
\hfill
\end{minipage}
\end{algorithm*}
\begin{figure*}[t]
    \centering
    \begin{subfigure}[t]{0.56\textwidth}
        \includegraphics{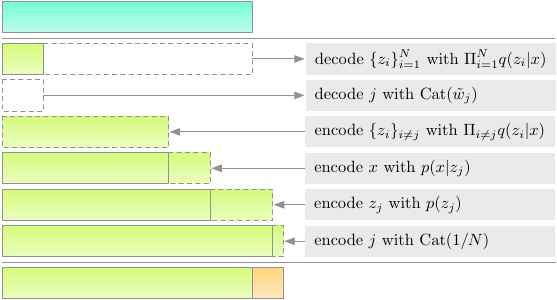}
        \caption{Bits-Back Importance Sampling (BB-IS)}
        \label{fig:bb_is}
    \end{subfigure}
    \hfill
    \begin{subfigure}[t]{0.43\textwidth}
        \includegraphics{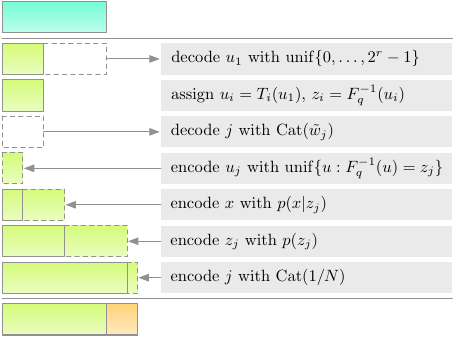}
        \caption{Bits-Back Coupled Importance Sampling (BB-CIS)}
        \label{fig:bb_cis}
    \end{subfigure}
    \caption{The initial bit cost of encoding a single symbol $x$ with IS-based coders is reduced from $\gO(N)$ to $\gO(\log N)$ by coupling the latents with shared randomness. Both of these coders achieve a net bitrate that approaches $-\log p(x)$ as $N \to \infty$.}\label{fig:bbis_bbcis}
\end{figure*}

\section{Monte Carlo Bits-Back Coding}
\label{sec:mcbits}
The net bitrate of bits-back is ideally the negative ELBO. This rate seems difficult to improve without finding a better $q$. However, the ELBO may be a loose bound on the marginal log-likelihood. Recent work in variational inference shows how to bridge the gap from the ELBO to the marginal log-likelihood with tighter variational bounds \citep[e.g.,][]{burda2015importance, domke2018importance}, motivating the question: \textit{can we derive bits-back coders from those tighter bounds and approach the cross-entropy?}

In this section we provide an affirmative answer to this question with a framework called Monte Carlo bits-back coding (McBits). We point out that the extended space constructions of Monte Carlo estimators can be reinterpreted as bits-back coders. One of our key contributions is deriving variants whose net bitrates improve over BB-ELBO, while being nearly as efficient with initial bits. We begin by motivating our framework with two worked examples. Implementation details are in Appendix \ref{appendix:mcbits_coders}.

\subsection{Bits-Back Importance Sampling}
\label{sec:bb_is}
The simplest of our McBits coders is based on importance sampling (IS). IS samples $N$ particles $z_i \sim q(z_i \given  x)$ i.i.d.\ and uses the {importance weights} $p(x, z_i)/q(z_i \given x)$ to estimate $p(x)$. The corresponding variational bound \citep[IWAE,][]{burda2015importance} is the log-average importance weight:
\begin{equation}
\label{eq:iwae}
       -\mathbb{E}_{\{z_i\}_{i=1}^N} \left[\log \left(\sum_{i=1}^N \frac{1}{N}\frac{p(x, z_i)}{q(z_i \given x)}\right) \right] \geq -\log p(x).
\end{equation}
IS provides a consistent estimator of $p(x)$. If the importance weights are bounded, the left-hand side of \eqref{eq:iwae} converges monotonically to $-\log p(x)$ \citep{burda2015importance}.

Surprisingly, \eqref{eq:iwae} is actually the evidence lower bound between a different model and a different approximate posterior on an extended space \cite{andrieu2010particle, cremer2017reinterpreting,domke2018importance}. In particular, consider an expanded latent space $\latentsymbolspace^N \times \intint{1}{N}$ that includes the configurations of the $N$ particles $\{z_i\}_{i=1}^N$ and an index $j \in \intint{1}{N}$. The left-hand side of \eqref{eq:iwae} can be re-written as the (negative) ELBO between a pair of distributions $P$ and $Q$ defined over this extended latent space, which are given in Alg. \ref{alg:is_extend_rep}. Briefly, given $x$, $Q$ samples $N$ particles $z_i \sim q(z_i \given x)$ i.i.d.\ and selects one of them by sampling an index $j$ with probability $\tilde{w}_j \propto p(x, z_j)/q(z_j \given x)$. The distribution $P$ pre-selects the special $j$th particle uniformly at random, samples its value $z_j \sim p(z_j)$ from the prior of the underlying model, and samples $x \sim p(x \given z_j)$ given $z_j$. The remaining $z_i \sim q(z_i \given x)$ for $i \neq j$ are sampled from the underlying approximate posterior. Because  $P$ and $Q$ use $q$ for all but the special $j$th particle, most of the terms in the difference of the log-mass functions cancel, and all that remains is \eqref{eq:iwae}. See Appendix \ref{appendix:bb_is}.

Once we identify the left-hand side of \eqref{eq:iwae} as a negative ELBO over the extended space, we can derive a bits-back scheme that achieves an expected net bitrate equal to \eqref{eq:iwae}. We call this the Bits-Back Importance Sampling (BB-IS) coder, and it is visualized in Fig. \ref{fig:bb_is}. To encode a symbol $x$, we first decode $N$ particles $z_i$ and the index $j$ with the $Q$ process by translating each `sample' to `decode'. Then we encode $x$, the particles $z_i$, and the index $j$ jointly with the $P$ process by translating each `sample' to `encode' in \emph{reverse order}. By reversing $P$ at encode time, we ensure that receiver decodes with $P$ in the right order.

Thus, ignoring the clean bits question, BB-IS's asymptotic net bitrate is close to the left-hand side of \eqref{eq:iwae}, which converges to the marginal log-likelihood \citep{burda2015importance}. Ultimately, as $N \to \infty$, it reaches the cross-entropy.

\subsection{Bits-Back Coupled Importance Sampling}
\label{sec:bb_cis}

Unfortunately, the BB-IS coder requires roughly $-  \log \tilde{w}_j - \sum_{i=1}^N \log q(z_i \given x) \in \gO(N)$ initial bits. The reason is that each decoded random variable needs to remove some bits from $m$. Can this be avoided? Here, we design Bits-Back Coupled Importance Sampling (BB-CIS), which achieves a net bitrate comparable to BB-IS while reducing the initial bit cost to $\gO(\log N)$. BB-CIS achieves this by decoding a \emph{single} common random number, which is shared by the $z_i$. The challenge is showing that a net bitrate comparable to BB-IS is still achievable under such a reparameterization.

\begin{figure}[t]
    \centering
    \includegraphics{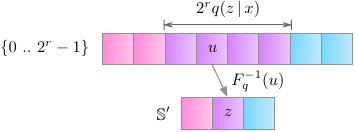}
    \caption{Mapping uniforms in $\intint{0}{2^r-1}$ to samples in $\latentsymbolspace$. Colors represent the subsets mapped to $z$ of size $2^r q(z \given x)$.}
    \label{fig:bb_cis_mapping}
\end{figure}

\begin{figure}[t]
    \centering
    \includegraphics{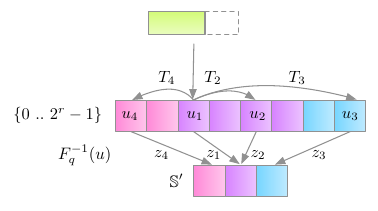}
    \caption{BB-CIS only needs to decode a single random uniform $u$ and then applies bijections $T_i$ to produce the uniforms $u_i$ underlying $z_i$. We set $T_1(u)=u$ by convention.}
    \label{fig:bb_cis_reparameterization}
\end{figure}

BB-CIS is based on a reparameterization of the particles $z_i$ as deterministic functions of coupled uniform random variables. The method is a discrete analog of the inverse CDF technique for simulating non-uniform random variates \citep{devroye2006nonuniform}.
Specifically, suppose that the latent space $\latentsymbolspace$ is totally ordered, and the probabilities of $q$ are approximated to an integer precision $r > 0$, i.e., $2^r q(z \given x)$ is an integer for all $z \in \latentsymbolspace$.
Define the function $F_q^{-1} : \intint{0}{2^r-1} \to \latentsymbolspace$,
\begin{equation}
    F_q^{-1}(u) = \arg \min \left\{z : \sum\nolimits_{z' \leq z} 2^r q(z' \given x) > u \right\}. \label{eq:gen_inv_cdf}
\end{equation}
$F^{-1}_q$ maps the uniform samples into samples from $q$. This is visualized in Fig. \ref{fig:bb_cis_mapping}.
A coupled set of particles $z_i$ with marginals $q(z_i \given x)$ can be simulated with a common random number by sampling a single uniform $u$ and setting $z_i = F_q^{-1}(T_i(u))$ for some functions $T_i : \intint{0}{2^r-1} \to \intint{0}{2^r-1}$, as in Fig. \ref{fig:bb_cis_reparameterization}. Intuitively, $T_i$ maps $u$ to the uniform $u_i$ underlying $z_i$. To ensure that $z_i$ have the same marginal, we require that $T_i$ are bijective functions. For example, $T_i$ can be defined as a fixed `shift' by integer $k_i$
, i.e.,  $T_i(u) = (u + k_i) \Mod{2^r}$ with inverse $T_i^{-1}(u) = (u - k_i) \Mod{2^r}$. We set $T_1(u) = u$ by convention.

\begin{figure*}[t!]
    \centering
    \begin{subfigure}{0.32\textwidth}
        \includegraphics[width=\textwidth]{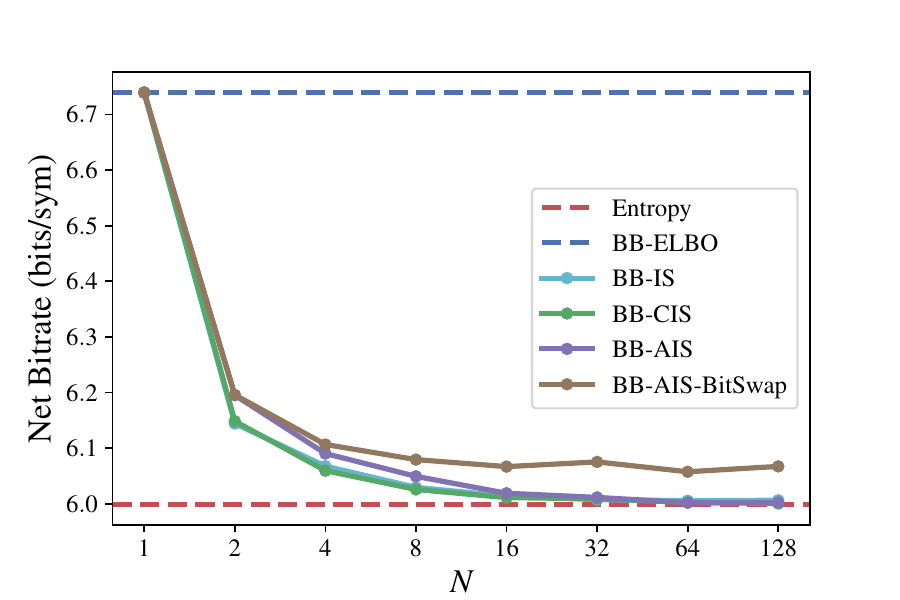}
        \caption{As $N \to \infty$, the net bitrate converges to the entropy for most coders on the toy mixture model.}
        \label{fig:convergence_plot_toy_mixture}
    \end{subfigure}
    \hfill
    \begin{subfigure}{0.32\textwidth}
        \includegraphics[width=\textwidth]{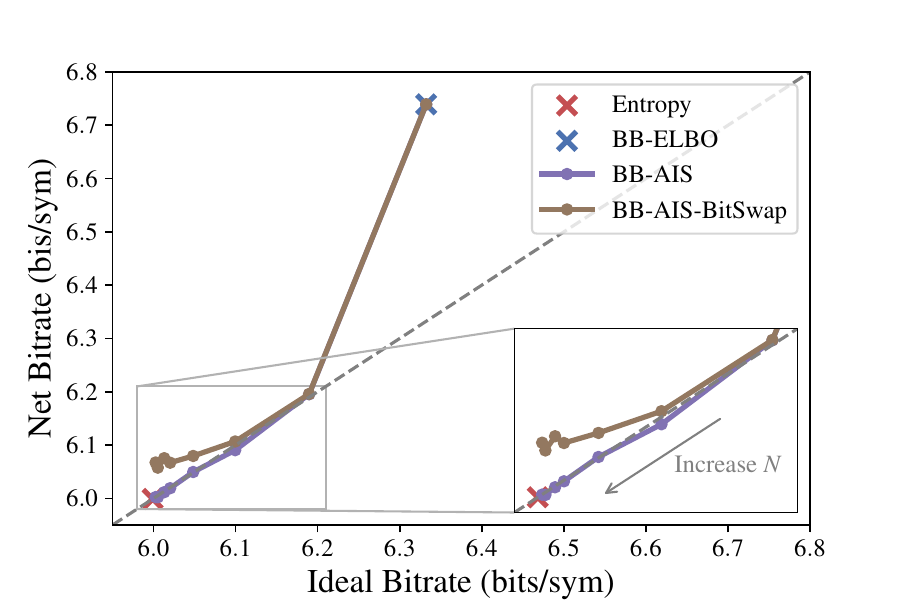}
        \caption{As $N \to \infty$, the net bitrate converges. Convergence to the ideal bitrate (dashed line) indicates clean bits.}
        \label{fig:cleanliness_plot_toy_mixture_only_ais}
    \end{subfigure}
    \hfill
    \begin{subfigure}{0.32\textwidth}
        \includegraphics[width=\textwidth]{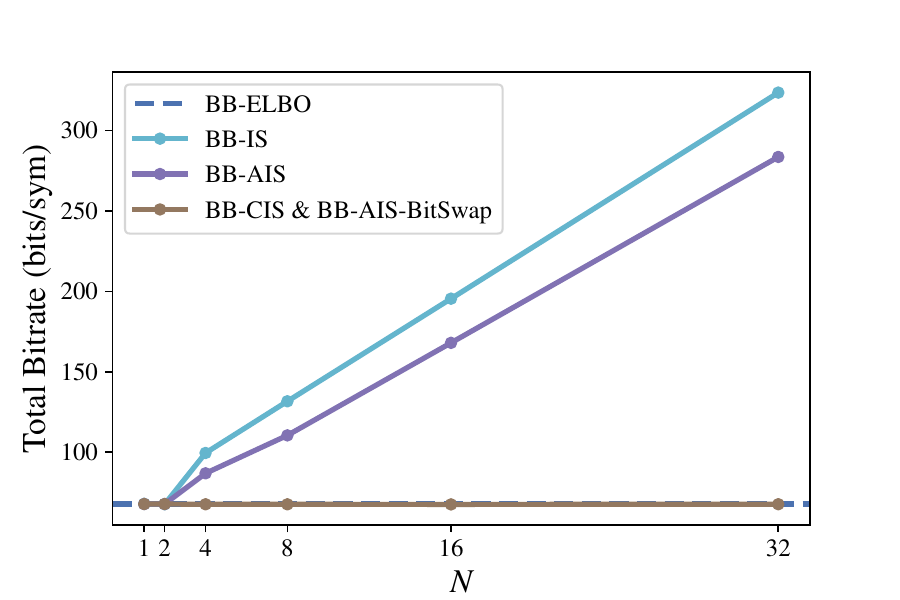}
        \caption{The initial bit cost (reflected in the total bitrate after the first symbol) is controlled by some McBits coders, but not others.}
        \label{fig:initial_bits_toy_mixture}
    \end{subfigure}
    \caption{The bitrate of McBits coders converges (in the number of particles $N$) to the entropy when using the data generating distribution. The initial bit cost of naive coders scales like $\gO(N)$, but coupled and BitSwap variants significantly reduce it. Bitrates are bits/sym.}
    \label{fig:toy_mixture_result}
\end{figure*}

BB-CIS uses these couplings in a latent decoding process that saves initial bits. It decodes $u_1$ from $m$ with $\uniform\intint{0}{2^r-1}$, sets $z_i = F_q^{-1}(T_i(u_1))$, and decodes the index $j$ of the special particle $z_j$ with $\cat(\tilde{w}_j)$. This reduces the initial bit cost to $r - \log \tilde{w}_j \in \gO(1) + \gO(\log N) = \gO(\log N)$. Note that the $\gO(\log N)$ term is for decoding the index $j$ which does not scale with latent dimension, thus the $\gO(1)$ term dominates for high dimensional latents. Also, in the ANS implementation, all compressed message lengths are rounded to a multiple of a specified precision (e.g., 16) which may mask small changes caused by the $\gO(\log N)$ term. Thus, in practice, BB-CIS demonstrates a nearly constant initial bits cost (Fig.  \ref{fig:initial_bits_toy_mixture}).

The coupled latent decoding process needs to be matched with an appropriate encoding process for $x$. Suppose that we finished encoding $x$, as with BB-IS, by encoding $\{z_i\}_{i \neq j}$ with $q(z_i \given x)$ i.i.d.\ and encoding $(x, z_j, j)$ with $p(x, z_j)/N$. The net bitrate would be
\begin{equation*}
    \expect_{u_1}\left[-r - \sum_{i = 1}^N \log q(z_i \given x) - \log \left(\sum_{i=1}^N \frac{1}{N}\frac{p(x, z_i)}{q(z_i \given x)}\right)\right].
\end{equation*}
This is clearly worse than BB-IS. The culprits are the $N-1$ latents that BB-IS pushes onto $m$, which is wasteful, because they are deterministically coupled. Fundamentally, the encoding process of BB-IS is not balanced with the initial bit savings of our coupled latent decoding process.

The solution is to design an encoding process over $\{z_i\}_{i =1}^N$, $\{u_i\}_{i = 1}^N$, and $x$, which exactly matches the initial bit savings of the coupled decoding process. The idea is to encode just $(u_j, x, z_j, j)$, which is enough information for the receiver to reconstruct all other variables. The key is to design the encoding for $u_j$. Encoding $u_j$ with a uniform on $\intint{0}{2^r-1}$ is unnecessarily wasteful, because $z_j$ restricts the range of $u_j$. It turns out, that the best we can do is to encode $u_j$ with $\uniform\{u : F^{-1}_q(u) = z_j\}$. Finally $(x, z_j, j)$ are encoded with $p(x, z_j)/N$. BB-CIS's encode is given in Fig. \ref{fig:bb_cis}.

BB-CIS has the following expected net bitrate:
\begin{equation}
    -\expect_{u_1}\left[\log\left(\sum_{i=1}^N \frac{1}{N}\frac{p(x, z_i)}{q(z_i \given x)}\right)\right].
\end{equation}
Thus, BB-CIS achieves a net bitrate comparable to BB-IS, but uses only $\gO(\log N)$ initial bits. We show this in Appendix \ref{appendix:bb_cis}. We also show that the BB-CIS net bitrate can be interpreted as a negative ELBO over an extended latent space, which accounts for the configurations of $z_i, u_i,$ and $j$. This extended space construction is novel, and we believe it may be useful for deriving other coupling schemes that reduce initial bit consumption. Common random numbers are classical tools in Monte Carlo methods; in bits-back they may serve as a tool for controlling initial bit cost.

The net bitrate of BB-CIS can fail to converge to $-\log p(x)$, if the $T_i$ are poorly chosen. In our experiments, we used fixed (but randomly chosen) shifts shared between the sender and receiver. This scheme converged as quickly as BB-IS in terms of net bitrate, but at a greatly reduced total bitrate. However, we point out that further work is required to explore more efficient bijections in terms of computation cost and compression performance.

\subsection{General Framework}

Monte Carlo bits-back coders generalize these two examples. They are bits-back coders built from extended latent space representations of Monte Carlo estimators of the marginal likelihood. Let $\hat{p}_N(x)$ be a positive unbiased Monte Carlo estimator of the marginal likelihood that can be simulated with $\gO(N)$ random variables, i.e., $\expect [\hat{p}_N(x)] = p(x)$.
Importance sampling is the quintessential example, $\hat{p}_N(x) = N^{-1}\sum_{i=1}^N p(x, z_i)/q(z_i\given x)$, but more efficient estimators of $p(x)$ can be built using techniques like annealed importance sampling \citep{neal2001annealed} or sequential Monte Carlo \citep{doucet2001introduction}.

A variational bound on the log-marginal likelihood can be derived from $\hat{p}_N(x)$ by Jensen's inequality,
\begin{equation}
    -\expect[\log \hat{p}_N(x)] \geq - \log p(x).
\end{equation}
If $\hat{p}_N(x)$ is strongly consistent in $N$ (as is the case with many of these estimators) and $-\log \hat{p}_N(x)$ satisfies a uniform integrability condition, then $-\expect[\log \hat{p}_N(x)] \to -\log p(x)$ \citep{maddison2017filtering}. This framework captures recent efforts on tighter variational bounds \citep{burda2015importance, maddison2017filtering, naesseth2018vsmc, le2018auto, domke2018importance, caterini2018hamiltonian}.

As with BB-IS and BB-CIS, the key step in the McBits framework is to identify an extended latent space representation of $\hat{p}_N(x)$. Let $\gZ \sim Q(\gZ \given x)$ be a set of random variables (often including those needed to compute $\hat{p}_N(x)$).
If there exists a `target' probability distribution $P(x, \mathcal{Z})$ over $x$ and $\gZ$ with marginal $p(x)$ such that
\begin{equation}
    \hat{p}_N(x)=\frac{P(x, \mathcal{Z})}{Q(\mathcal{Z} \given x)},
\end{equation}
then the McBits coder, which decodes $\gZ$ with $Q(\gZ \given x)$ and encodes $(x, \gZ)$ with $P(x, \gZ)$, will achieve a net bitrate of $-\log \hat{p}_N(x)$. In particular, if the log estimator converges in expectation to the log-marginal likelihood and we ignore the dirty bits issue, then $\kl{Q(\gZ \given x)}{P(\gZ \given x)} \to 0$ and the McBits coder will achieve a net bitrate of $H(p_d, p)$.

The challenge is to identify $\gZ$, $Q$, and $P$. While Monte Carlo estimators of $p(x)$ get quite elaborate, many of them admit such extended latent space representations \citep[e.g.,][]{neal2001annealed, andrieu2010particle, finke2015extended, domke2018importance}. These constructions are techniques for proving the unbiasedness of the estimators; one of our contributions is to demonstrate that they can become efficient bits-back schemes. In Appendix \ref{appendix:mcbits_coders}, we provide pseudocode and details for all McBits coders.

\paragraph{Bits-Back Annealed Importance Sampling (BB-AIS)}
Annealed importance sampling (AIS) is a generalization of importance sampling, which introduces a path of $N$ intermediate distributions between the base distribution $q(z \given x)$ and the unnormalized posterior $p(z \given x)$. AIS samples a sequence of latents by iteratively applying MCMC transition kernels that leave each intermediate distributions invariant. By bridging the gap between $q(z \given x)$ and $p(z \given x)$, AIS's estimate of $p(x)$ typically converges faster than importance sampling \citep{neal2001annealed}.
The corresponding McBits coder, BB-AIS, requires $\gO(N)$ initial bits, but this can be addressed with the BitSwap trick \citep{kingma2019bit}, which we call BB-AIS-BitSwap. Another issue is that the intermediate distributions are usually not factorized, which makes it challenging to work with high-dimensional $z$.

\paragraph{Bits-Back Sequential Monte Carlo (BB-SMC)}
Sequential Monte Carlo (SMC) is a particle filtering method that combines importance sampling with resampling. Its estimate of $p(x)$ typically converges faster than importance sampling for time series models \cite{cerou2011nonasymptotic, berard2014lognormal}. SMC maintains a population of particles. At each time step, the particles independently sample an extension from the proposal distribution. Then the whole population is resampled with probabilities in proportion to importance weights. The corresponding McBits coder, BB-SMC, requires $\gO(TN)$ initial bits, where $T$ is the length of the time series. We introduce a coupled variant, BB-CSMC, in the appendix that reduces this to $\gO (T \log N)$.

\begin{figure}[ht]
    \vspace{-.5\baselineskip}
    \centering
    \includegraphics[width=0.36\textwidth]{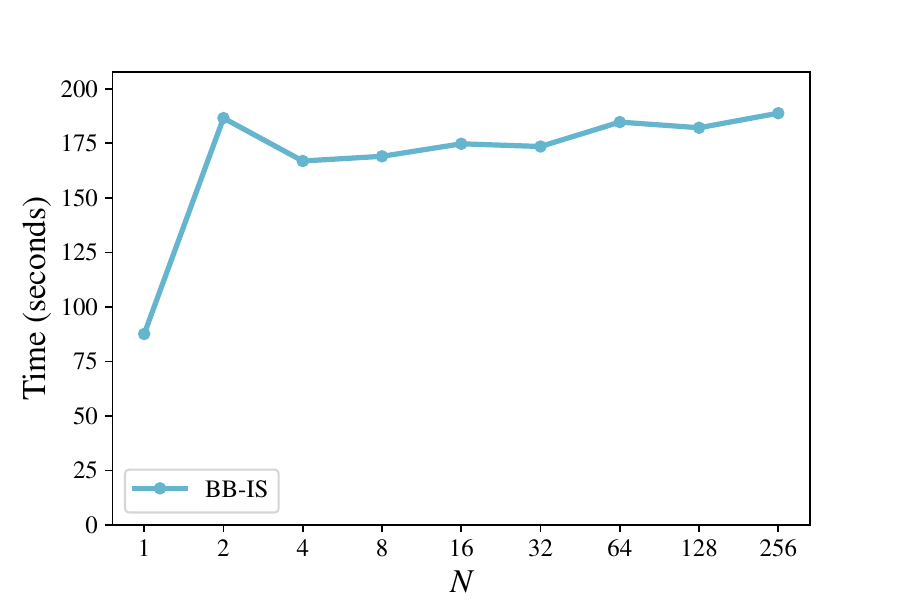}
    \caption{BB-IS total encode + decode time for binarized MNIST scales very well with $N$ using a parallel implementation of ANS. BB-ELBO ($N=1$) uses a simpler code-base, which runs faster. The experiment was run on a Tesla P100 GPU with 12GB of memory, together with an Intel Xeon Silver 4110 CPU at 2.10GHz.}
    \label{fig:time_plot}
    \vspace{-\baselineskip}
\end{figure}

\paragraph{Computational Cost}
All of our coders require $\gO(N)$ computational cost, but the IS- and SMC-based coders are amenable to parallelization over particles. In particular, we implemented an end-to-end parallelized version of BB-IS based on the JAX framework \citep{bradbury2018} and benchmarked on compressing the binarized MNIST dataset with one-layer VAE model. As shown in Fig. \ref{fig:time_plot}, the computation time scales sublinearly with the number of particles, which demonstrates the potential practicality of our method, in some settings, with hundreds of particles. Detailed discussion is in Appendix \ref{appendix:compcost}.

 \section{Related Work}

Recent work on learned compressors considers two classes of deep generative models: models based on normalizing flows
\cite{rippel2013high,dinh2014nice, dinh2016density, rezende2015variational} and
deep latent variable models \cite{kingma2013auto, rezende2014stochastic}.
Flow models are generally computationally expensive, and optimizing a function with discrete domain and
codomain using gradient descent can be cumbersome. Despite these issues,
flow models are the state of the art for lossless
compression of small images \cite{van2020idf++}.

`Bits-back' was originally meant to provide an information theoretic basis for
approximate Bayesian inference methods \cite{wallace1990,hinton1993keeping,frey1997efficient,frey1998graphical}. \citet{townsend2019practical} showed
that the idea can lead directly to a practical compression algorithm for latent variable models. Follow up work reduced the initial bit cost for hierarchical models \cite{kingma2019bit}, and extended it to large scale
models and larger images \cite{townsend2020hilloc}. For lossy compression, the variational autoencoder framework is a natural fit for training transform coders \cite{johnston2019computationally,
balle2016end, balle2018variational, minnen2018joint, yang2020improving}.

Relative entropy coding \citep[REC,][]{havasi2018minimal, flamich2020rec} is an alternative coding scheme for latent variable models that seeks to address the initial bits overhead in bits-back schemes. However, practical REC implementations require a particular reparameterization of the latent space, and it is unclear whether the REC latent structure is compatible with the extended latents in McBits.

 \section{Experiments}

We studied the empirical properties and performance of our McBits coders on both synthetic data and practical image and music piece compression tasks.
Many of our experiments used continuous latent variable models and we adopted the maximum entropy quantization in \citet{townsend2019practical} to discretize the latents. We sometimes evaluated the \emph{ideal bitrate}, which for each coder is the corresponding variational bound estimated with pseudorandom numbers. For continuous latent variable models, the ideal bitrate does not account for quantization. We rename BB-ANS to BB-ELBO. $N$ refers to the number of intermediate distributions for BB-AIS. Details are in Appendix \ref{appendix:experimental_details}. Our implementation is available at \url{https://github.com/ryoungj/mcbits}.

\subsection{Lossless Compression on Synthetic Data}

We assessed the convergence properties, impact of dirty bits, and initial bit cost of our McBits coders on synthetic data.
First, a dataset of 5000 symbols was generated i.i.d.\ from a mixture model with alphabet sizes 64 and 256 for the observations and latents, respectively. BB-ELBO, BB-IS, BB-CIS, and BB-AIS were evaluated using the true data generating distribution with a uniform approximate posterior, ensuring a large mismatch with the true posterior.
For BB-AIS, a Metropolis--Hastings kernel with a uniform proposal was used. The bijective operators of BB-CIS applied randomly selected, but fixed, shifts to the sampled uniform.

The net bitrates of BB-IS, BB-CIS, and BB-AIS converged to the entropy (optimal rate) as the number of particles increased. This is shown in \ref{fig:convergence_plot_toy_mixture}.
The indistinguishable gap between BB-CIS and BB-IS illustrates that particle coupling did not lead to a deterioration of net bitrate. BB-AIS-BitSwap did not converge, likely due to the dirty bits issue.

We measured the impact of dirty bits by plotting \textit{ideal} versus \textit{net} bitrates in Fig.  \ref{fig:cleanliness_plot_toy_mixture_only_ais}. The deviation of any point to the dashed line indicates the severity of dirty bits. Interestingly, most of our McBits coders appeared to `clean' the bitstream as $N$ increased, i.e. the net bitrate converged to the entropy, as shown in Fig. \ref{fig:cleanliness_plot_toy_mixture_only_ais} for BB-AIS and in Fig. \ref{fig:cleanliness_plot_toy_mixture_all} in Appendix \ref{appendix:lossless_toy} for all coders. Only BB-AIS-BitSwap did \emph{not} clean the bitstream (Fig. \ref{fig:cleanliness_plot_toy_mixture_only_ais}), indicating that the order of operations has a significant impact on the cleanliness of McBits coders.

We quantified the initial bits cost by computing the \textit{total} bitrate after the first symbol. As shown in Fig.  \ref{fig:initial_bits_toy_mixture}, it increased linearly with $N$ for BB-IS and BB-AIS, but remained fixed for BB-CIS and BB-AIS-BitSwap.

\begin{figure}[thb]
    \centering
    \includegraphics[width=0.34\textwidth]{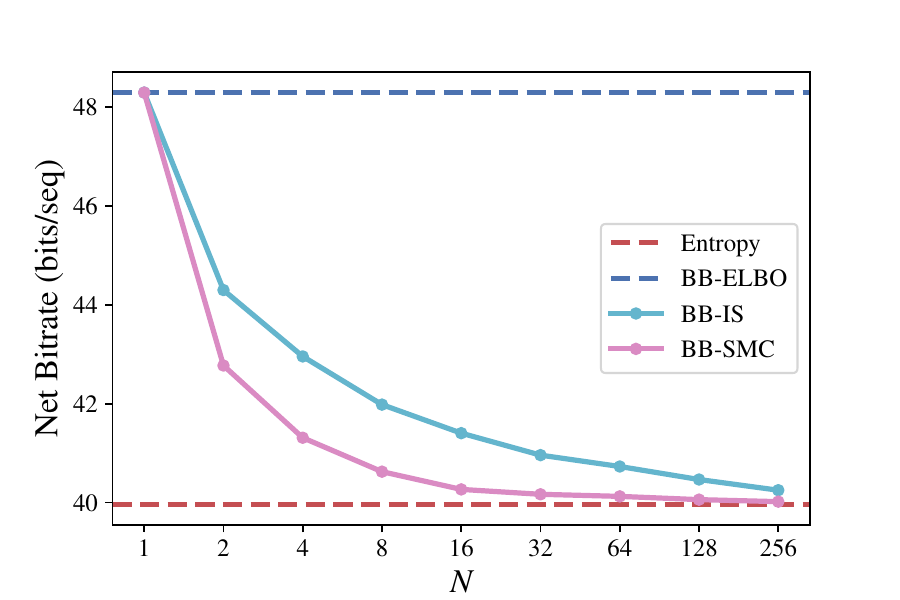}
    \vspace{-0.5\baselineskip}
    \caption{As $N \rightarrow \infty$, the net bitrates of BB-IS and BB-SMC converge to the entropy on the toy HMM, but BB-SMC converges much faster.}
    \label{fig:convergence_plot_toy_hmm}
\end{figure}

Our second experiment was with a dataset of 5000 symbol subsequences generated i.i.d.\ from a small hidden Markov model (HMM). We used 10 timesteps with alphabet sizes 16 and 32 for the observations and latents, respectively. BB-ELBO, BB-IS, and BB-SMC were evaluated using the true data generating distribution and a uniform approximate posterior. The net bitrates of BB-IS and BB-SMC converged to the entropy, but BB-SMC converged much faster, illustrating the effectiveness of resampling particles for compressing sequential data (Fig.  \ref{fig:convergence_plot_toy_hmm}).

\subsection{Lossless Compression on Images}

We benchmarked the performance of BB-IS and BB-CIS on the standard train-test splits of two datasets: an alphanumeric extension of MNIST called EMNIST \citep{cohen2017emnist}, and CIFAR-10 \citep{krizhevsky2009learning}. EMNIST was dynamically binarized following \citet{salakhutdinov2008quantitative}, and a VAE with 1 stochastic layer was used as in \citet{burda2015importance}.
For CIFAR-10, we used VQ-VAE \citep{oord2017neural} with discrete latent variables and trained with continuous relaxations \citep{sonderby2017continuous}.

\begin{table}[tb]
\vspace{-0.25\baselineskip}
\caption{BB-IS performs better on models that were trained with the same number of particles. The net bitrates (bits/dim) of BB-IS on EMNIST-MNIST and CIFAR-10 test sets. $N=50$ for EMNIST-MNIST and $N=10$ for CIFAR-10.}
\label{table:vae_training_method}
\begin{center}
\begin{small}
\begin{tabular}{lcccc}
\toprule
  & \multicolumn{2}{c}{MNIST} &   \multicolumn{2}{c}{CIFAR-10} \\
 \cmidrule(lr){2-3} \cmidrule(lr){4-5}
 & ELBO & IWAE &  ELBO & IWAE \\
\midrule
BB-ELBO & 0.236 &0.236 & 4.898 &  4.898 \\
BB-IS (5) & 0.232 & 0.231 &  4.866 &4.827 \\
BB-IS ($N$) & 0.230 & \textbf{0.228} & 4.857 &  \textbf{4.810} \\
\midrule
Savings & 2.5\% & 3.4\% &  0.8\% & 1.8\% \\
\bottomrule
\end{tabular}
\end{small}
\end{center}
\end{table}

\begin{table}[t]
\caption{BB-IS leads to more improved compression rates in out-of-distribution compression settings. The net bitrates (bits/dim) of BB-IS on EMNIST test sets using a VAE.}
\label{table:vae_transfer}
\begin{center}
\begin{small}
\begin{tabular}{lcccc}
\toprule
\multicolumn{1}{c}{Trained on} & \multicolumn{2}{c}{MNIST} & \multicolumn{2}{c}{Letters} \\ \cmidrule(lr){1-1} \cmidrule(lr){2-3} \cmidrule(lr){4-5}
\multicolumn{1}{c}{Compressing} & MNIST & Letters  & MNIST & Letters\\
\midrule
BB-ELBO & 0.236 & 0.310  & 0.257 & 0.250 \\
BB-IS (5) & 0.231 & 0.289 & 0.249 & 0.243 \\
BB-IS (50) & 0.228 & 0.280 & 0.244 & 0.239\\
\midrule
Savings & 3.4\% & \textbf{9.7\%} & \textbf{5.1\%} & 4.4\% \\
\bottomrule
\end{tabular}
\end{small}
\end{center}
\end{table}

The variational bounds used to train the VAEs had an impact on compression performance. When using BB-IS with a model trained on the IWAE objective, equalizing the number of particles during compression and training resulted in better rates than BB-IS with an ELBO-trained VAE (Table \ref{table:vae_training_method}). Therefore, we always use our McBits coders with models trained on the corresponding variational bound.

We assessed BB-IS in an out-of-distribution (OOD) compression setting. We trained models on standard EMNIST-Letters and EMNIST-MNIST splits and evaluated compression performance on the test sets. BB-IS achieved greater rate savings than BB-ELBO when transferred to OOD data (Table \ref{table:vae_transfer}). This illustrates that BB-IS may be particularly useful in more practical compression settings where the data distribution is different from that of the training data.

\begin{table}[t]
\vspace{-0.5\baselineskip}
\caption{BB-CIS achieves the best total bitrates compared to baselines on EMNIST test sets.}
\label{table:emnist_baseline_compare}
\begin{center}
\begin{small}
\begin{tabular}{ccc}
\toprule
    Method         &   MNIST    &   Letters  \\
\midrule
    PNG     &   0.819    &   0.900    \\
    WebP    &   0.464    &   0.533    \\
    gzip    &   0.423    &   0.413    \\
    lzma    &   0.383    &   0.369    \\
    bz2     &   0.375    &   0.364    \\
\midrule
    \multicolumn{1}{l}{BB-ELBO}             &   0.236          &    0.250   \\
     \multicolumn{1}{l}{BB-ELBO-IF (50)}     &   0.233          &    0.246  \\
     \multicolumn{1}{l}{BB-IS (50)}          &   0.230          &    0.241   \\
     \multicolumn{1}{l}{BB-CIS (50)}         &   \textbf{0.228} & \textbf{0.239}  \\
\bottomrule
\end{tabular}
\end{small}
\end{center}
\end{table}

Finally, we compared BB-IS and BB-CIS to other benchmark lossless compression schemes by measuring the \textit{total} bitrates on EMNIST test sets (without transferring).
We also compared with amortized-iterative inference, \citep{yang2020improving} that optimizes the ELBO objective over local variational parameters for each data example at the compression stage. To roughly match the computation budget, the number of optimization steps was set to 50 and this method is denoted as BB-ELBO-IF (50). Both BB-IS and BB-CIS outperformed all other baselines on both test sets, and BB-CIS was better than BB-IS in terms of total bitrate since it effectively reduces the initial bit cost. Additional results can be found in in Appendix \ref{appendix:lossless_image:additional}.

\subsection{Lossless Compression on Sequential Data}
\begin{table}[t]
\vspace{-0.5\baselineskip}
\caption{BB-SMC achieves the best net bitrates (bits/timestep) on all piano roll test sets.}
\label{table:seqential_data_bb_smc_result}
\begin{center}
\begin{small}
\begin{tabular}{lcccc}
\toprule
             &   Musedata    &   Nott.   &   JSB     &   Piano.     \\
\midrule
    BB-ELBO     &    10.66   &    5.87   &   12.53    &    11.43   \\
    BB-IS (4)     &     10.66      &  4.86     &   12.03    &   11.38 \\
    BB-SMC (4)     &   \textbf{9.58}    &   \textbf{4.76}    &   \textbf{10.92}    &   \textbf{11.20}   \\
\midrule
Savings & 10.1\% & 18.9\% & 12.8\% & 2.0\% \\
\bottomrule
\end{tabular}
\end{small}
\end{center}
\vspace{-1\baselineskip}
\end{table}
We quantified the performance of BB-SMC on sequential data compression tasks with 4 polyphonic music datasets: Nottingham, JSB, MuseData, and Piano-midi.de \citep{boulanger2012modeling}. All datasets were composed of sequences of binary 88-dimensional vectors representing active notes.
The sequence lengths were very imbalanced, so we chunked the datasets to sequences with maximum length of 100.
We trained variational recurrent neural networks \citep[VRNN,][]{chung2015recurrent} on these chunked datasets using code from \citet{maddison2017filtering}.
For each dataset, 3 VRNN models were trained with the ELBO, IWAE and FIVO objectives with 4 particles and were used with their corresponding coders for compression.

We compared the \textit{net} bitrates of all coders for compressing each test set in Table \ref{table:seqential_data_bb_smc_result}. BB-SMC clearly outperformed BB-ELBO and BB-IS  with the same number of particles on all datasets. We include the comparison with some benchmark lossless compression schemes in Table \ref{table:seqential_data_baselines_compare} in Appendix \ref{appendix:lossless_sequential:additional}.

\subsection{Lossy Compression on Images}

Current state-of-the-art lossy image compressors use hierarchical latent VAEs with quantized latents that are losslessly compressed with hyperlatents \citep{balle2016end,balle2018variational, minnen2018joint}.
\citet{yang2020improving} observed that the marginalization gap of jointly compressing the latent and the hyperlatent can be bridged by bits-back. Thus, our McBits coders can be used to further reduce the gap.

We experimented on a simplified setting where we used the binarized EMNIST datasets and a modification of the VAE model with 2 stochastic layers in \citet{burda2015importance}. The major modifications were the following. The distributions over the 1st stochastic layer (latent) were modified to support quantization in a manner similar to \cite{balle2018variational}.
We trained the model on a relaxed rate-distortion objective with a hyperparameter $\lambda$ controlling the trade-off, where the distortion term was the Bernoulli negative log likelihood and the rate term was the negative ELBO or IWAE that only marginalized over the 2nd stochastic layer (hyperlatent).
Details are in Appendix \ref{appendix:lossy_image}.

\begin{figure}[t]
    \vspace{-0.5\baselineskip}
    \centering
    \includegraphics[width=0.34\textwidth]{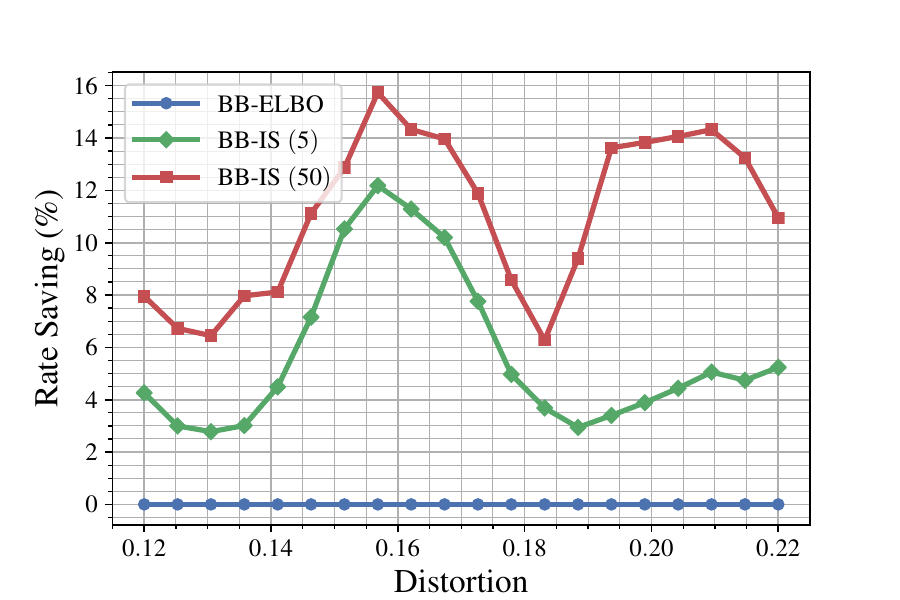}
    \vspace{-.5\baselineskip}
    \caption{The rate saving curve for lossy compression on the EMNIST-MNIST test set. We measure the net bitrate savings ($\%$) relative to BB-ELBO for fixed distortion values.}
    \label{fig:lossy_mnist_bb_is_rate_saving_curve}
\end{figure}

We evaluated the \textit{net} bitrate savings of BB-IS compared to BB-ELBO on the EMNIST-MNIST test set with different $\lambda$ values, as in Fig. \ref{fig:lossy_mnist_bb_is_rate_saving_curve}. We found that BB-IS achieved more than $15 \%$ rate savings in some setups, see also the rate-distortion curves in Appendix \ref{appendix:lossy_image:additional}. The performance can be further improved by applying amortized-iterative inference, which is included in Appendix \ref{appendix:lossy_image:additional}. We also implemented these experiments for the model in \cite{balle2018variational}, but did not observe significant improvements. This may be due to the specific and complex model architecture.

 \section{Conclusion}

We showed that extended state space representations of Monte Carlo estimators of the marginal likelihood can be transformed into bits-back schemes that asymptotically remove the $\KL$ gap. In our toy experiments, our coders were `self-cleaning' in the sense that they reduced the dirty bits gap. In our transfer experiments, our coders had a larger impact on compression rates when compressing out-of-distribution data. Finally, we demonstrated that the initial bit cost incurred by naive variants can be controlled by coupling techniques. We believe these coupling techniques may be of value in other settings to reduce initial bit costs.

\section*{Acknowledgements}

We thank Yann Dubois and Guodong Zhang for feedback on an early draft. We thank Roger Grosse and David Duvenaud for thoughtful comments. Arnaud Doucet acknowledges support from EPSRC grants EP/R018561/1 and EP/R034710/1. This research is supported by the Vector Scholarship in Artificial Intelligence. Resources used in preparing this research were provided, in part, by the Province of Ontario, the Government of Canada through CIFAR, and companies sponsoring the Vector Institute.

\bibliography{refs}
\bibliographystyle{icml2021}

\onecolumn

\appendix
\section{Monte Carlo Bits-Back Coders}
\label{appendix:mcbits_coders}

\subsection{Notation}

\begin{itemize}
    \item $\indicate{A}$ is the indicator function of the event $A$, i.e., $\indicate{A} = 1$ if $A$ holds and $0$ otherwise.
    \item $\sP(A)$ is the probability of the event $A$.
\end{itemize}

\subsection{Assumptions}

\begin{itemize}
    \item We assume that, if $q(z \given x) > 0$, then $p(x,  z) > 0$.
\end{itemize}

\subsection{General Framework}

\begin{minipage}[tbh]{0.49\textwidth}
\begin{algorithm}[H]
\caption{Encode Procedure of McBits Coders}
\label{alg:mcbits_encode}
\Procedure{\Encode{symbol $x$, message $m$}}{
    decode $\mathcal{Z}$ with $Q(\mathcal{Z} \given x)$\\
    encode $x$ and $\mathcal{Z}$ with $P(x, \mathcal{Z})$\\
    \KwRet $m'$
}
\end{algorithm}
\end{minipage}
\begin{minipage}[tbh]{0.49\textwidth}
\begin{algorithm}[H]
\caption{Decode Procedure of McBits Coders}
\label{alg:mcbits_decode}
\Procedure{\Decode{message $m$}}{
    decode $x$ and $\mathcal{Z}$ with $P(x, \mathcal{Z})$ \\
    encode $\mathcal{Z}$ with $Q(\mathcal{Z} \given x)$ \\
    \KwRet $x$, $m'$
}
\end{algorithm}
\end{minipage}
As discussed in the main body, given the extended space representation of an unbiased estimator of the marginal likelihood, we can derive the general procedure of McBits coders as Alg. \ref{alg:mcbits_encode}. The decode procedure could be easily derived from the encode procedure by simply reverse the order and switch `encode' with `decode', as in Alg. \ref{alg:mcbits_decode}. Therefore, we do not tediously present the decode procedure for each McBits coder in the following subsections.

\subsection{Bits-Back Importance Sampling (BB-IS)}
\label{appendix:bb_is}
Importance sampling (IS) samples $N$ particles $z_i \sim q(z_i \given  x)$ i.i.d. and uses the average importance weight to estimate $p(x)$. The corresponding variational bound \citep[IWAE,][]{burda2015importance} is the log-average importance weight which is given in \eqref{eq:iwae}. For simplicity, we denote importance weight as $w_i=p(x,z_i)/q(z_i\given x)$ and normalized importance weight as $\tilde{w}_i = w_i / \sum_i w_i$. The basic importance sampling estimator of $p(x)$ is given by
\begin{equation}
    \sum_{i=1}^N \frac{w_i }{N}= \sum_{i=1}^N \frac{1}{N}\frac{p(x,z_i)}{q(z_i\given x)}.
\end{equation}

\paragraph{Extended latent space representation} The extended latent space representation of the importance sampling estimator is presented in Alg. \ref{alg:is_extend_rep}, from which we can derive the proposal and target distribution as followed:
\begin{align}
    & Q(\{z_i\}_{i=1}^N, j \given x) =  \tilde{w}_j \prod_{i = 1}^N q(z_i \given x) \\
    & P(x, \{z_i\}_{i=1}^N, j) = \frac{1}{N} p(x, z_{j})\prod_{i \neq j} q(z_i \given x).
\end{align}
Now note,
\begin{align}
    \frac{P(x, \{z_i\}_{i=1}^N, j)}{Q(\{z_i\}_{i=1}^N, j \given x)} = \frac{1}{N} \frac{\sum_{i=1}^N w_i}{w_j} \frac{p(x, z_j)}{q(z_j \given x)} =  \sum_{i=1}^N\frac{w_i}{N},
\end{align}
which exactly gives us the IS estimator.

\begin{algorithm}[tbh]
\caption{Encode Procedure of BB-IS}
\label{alg:bb_is_encode}
\Procedure{\Encode{symbol $x$, message $m$}}{
    decode $\{z_i\}_{i=1}^N  $ with $  \prod_{i=1}^N q(z_i \given x)$ \\
    decode $j  $ with $  \cat\left(\tilde{w}_j\right)$ \\
    encode $\{z_i\}_{i \neq j}  $ with $  \prod_{i \neq j}q(z_i \given x)$ \\
    encode $x  $ with $  p(x \given z_j)$ \\
    encode $z_j  $ with $  p(z_j)$ \\
    encode $j  $ with $  \cat(1/N)$ \\
    \KwRet $m'$
}
\end{algorithm}

\paragraph{BB-IS Coder} Based on the extended latent space representation, the Bits-Back Importance Sampling (BB-IS) coder is derived in Alg. \ref{alg:bb_is_encode} and visualized in Fig. \ref{fig:bb_is}. The expected net bit length for encoding a symbol $x$ is:
\begin{align}
 - \expect_{\{z_i\}_{i=1}^N, j}\left[\log \frac{P(x, \{z_i\}_{i=1}^N, j)}{Q(\{z_i\}_{i=1}^N, j \given x)} \right] =  - \expect_{\{z_i\}_{i=1}^N}\left[ \log \left(\sum_{i=1}^N\frac{w_i}{N} \right) \right].
\end{align}
Ignoring the dirty bits issue, i.e., $z_i \sim q(z_i \given x)$ i.i.d., the expected net bit length exactly achieves the negative IWAE bound.

\subsection{Bits-Back Coupled Importance Sampling}
\label{appendix:bb_cis}

The basic idea behind Bits-Back Coupled Importance Sampling is to couple the randomness that generates the particles $z_i$. We assume that $q(z \given x)$ has been discretized to precision $r$. In particular, we assume that for all $z \in \latentsymbolspace$
\begin{equation}
    q(z \given x) = \frac{q_z}{2^r},
\end{equation}
where $q_z$ is an integer. Note that this assumption is also required for ANS, so this is not an additional assumption. Assume that $\latentsymbolspace$ is totally ordered (any ordering works) and $u \in \intint{0}{2^r-1}$. Define the unnormalized cumulative distribution function $F_q$ of $q$ as well as the following related objects:
\begin{align}
    F_q(z) &= \sum_{z' \leq z} q_{z'}\\
    F_q^{-1}(u) &= \arg\min \{z \in \latentsymbolspace : F_q(z) > u\}\\
    U(z) &= \{u : F_q^{-1}(u) = z\}.
\end{align}
Notice that $|U(z)| = q_z$. Thus, if $u \sim \uniform \intint{0}{2^r-1}$, then
\begin{align*}
    \sP\left(F_q^{-1}(u) = z\right) &= \sP\left( u \in U(z) \right)\\
    &= \frac{|U(z)|}{2^r}\\
    &= \frac{q_z}{2^r}.
\end{align*}
Therefore we can reparameterize the sampling process of $z$ as uniform sampling over $\intint{0}{2^r-1}$ followed by the deterministic mapping $F_q^{-1}$. This is a classical approach in non-uniform random variate generation, specialized to this discrete setting. This is visualized in Fig. \ref{fig:bb_cis_mapping}.

Coupled importance sampling (CIS) samples the latent variables by coupling their underlying uniforms. Let $T_i : \intint{0}{2^r-1} \to \intint{0}{2^r-1}$ be bijective functions. If $u \sim \uniform\intint{0}{2^r-1}$, then
\begin{align*}
    \sP\left(T_i(u) = u'\right) &= \sP\left(u = T_i^{-1}(u')\right)\\
    &= \frac{\indicate{T_i^{-1}(u') \in \intint{0}{2^r-1}}}{2^r}\\
    &= \frac{1}{2^r}.
\end{align*}
Thus, if $u \sim \uniform\intint{0}{2^r-1}$, then $T_i(u) \overset{d}{=} u$ and $F_{q}^{-1}(T_i(u)) \sim q(z \given x)$. For example, simple bijective operators $T_i$ can be defined as applying fixed ``sampling shift'' $\bar{u}_i \in \{0, 1, \dots, 2^r-1\}$ to $u$:
\begin{align*}
    T_i(u) &= (u+\bar{u}_i) \Mod{2^r}\\
    T_i^{-1}(u) &= (u-\bar{u}_i) \Mod{2^r}.
\end{align*}
For simplicity, we define $T_1$ to employ the zero shift $\bar{u}_1=0$, i.e., $T_1(u) = u$. We now have the definitions that we need to analyze the extended space construction for coupled importance sampling. Let $u_1 \sim \uniform\intint{0}{2^-1}$. As with IS, we denote importance weight as $w_i=p(x, F_q^{-1}(T_i(u_1)))/q(F_q^{-1}(T_i(u_1)) \given x)$ and the normalized importance weight as $\tilde{w}_i = w_i / \sum_i w_i$. The coupled importance sampling estimator of $p(x)$ is given

\begin{equation}
    \sum_{i=1}^N \frac{w_i }{N}= \sum_{i=1}^N \frac{1}{N}\frac{p(x,F_q^{-1}(T_i(u_1)))}{q(F_q^{-1}(T_i(u_1))\given x)}.
\end{equation}

\begin{algorithm}[t]
\caption{Extended Latent Space Representation of Coupled Importance Sampling}
\label{alg:cis_extend_rep}
\begin{minipage}[t]{0.46\textwidth}
\Process{$Q(\gZ \given x)$}{
    sample $u_1 \sim \uniform\{0, \ldots, 2^r - 1\}$ \\
    \For{$i = 1,\dots, N$}{
        assign $u_i = T_i(u_1)$\\
        assign $z_i = F_q^{-1}(u_i)$
    }
    compute $\tilde{w}_i \propto p(x, z_i) / q(z_i \given x)$\\
    sample $j \sim \cat(\tilde{w}_i)$ \\
    \KwRet $\{z_i\}_{i=1}^N, \{u_i\}_{i=1}^N, j$
}
\end{minipage}
\begin{minipage}[t]{0.46\textwidth}
\Process{$P(x, \gZ)$}{
    sample $j \sim \cat(1/N)$ \\
    sample $z_j \sim p(z_j)$\\
    sample $x \sim p(x | z_j)$ \\
    sample $u_j \sim \uniform \{u : F_q^{-1}(u) = z_j\}$ \\
    \For{$i \neq j$}{
        assign $u_i = T_i(T_j^{-1}(u_j))$\\
        assign $z_i = F_q^{-1}(u_i)$
    }
    \KwRet $x, \{z_i\}_{i=1}^N, \{u_i\}_{i=1}^N, j$
}
\end{minipage}
\hfill\end{algorithm}
\paragraph{Extended latent space representation} The extended latent space representations of the coupled importance sampling estimator is presented in Alg. \ref{alg:cis_extend_rep}. The $Q$ and $P$ processes have the following probability mass functions. For convenience, let $\sU = \intint{0}{2^r-1}$ and $z_i = F_q^{-1}(T_i(u_1))$.
\begin{align}
    Q(\{z_i\}_{i=1}^N, \{u_i\}_{i=1}^N, j \given x) &= \frac{\indicate{u_1 \in \sU}}{2^r} \prod_{i = 2}^N \indicate{T_i(u_1) = u_i}  \prod_{i = 1}^N  \indicate{F_q^{-1}(u_i) = z_i} \frac{w_j}{\sum_{i=1}^N w_i}\\
    P(x, \{z_i\}_{i=1}^N, \{u_i\}_{i=1}^N, j) &= \frac{1}{N} p(x, z_{j}) \frac{\indicate{u_j \in U(z_j)}}{2^r q(z_j \given x)} \prod_{i \neq j} \indicate{T_i\left(T_j^{-1}(u_j)\right) = u_i}   \indicate{F_q^{-1}(u_i) = z_i},
\end{align}
where we used the fact that $q_{z_j} = 2^r q(z_j \given x) =  \given U(z_j) \given $. Because each $T_i$ is a bijection, we have that the range of $T_i(u_1)$ is all of $\sU$, as $u_1$ ranges over $\sU$. Thus,
\begin{equation}
    \frac{\indicate{u_1 \in \sU}}{2^r} \prod_{i = 2}^N \indicate{T_i\left(u_1\right) = u_i}  = \frac{\indicate{u_j \in \sU}}{2^r} \prod_{i \neq j} \indicate{T_i\left(T_j^{-1}(u_j)\right) = u_i}.
\end{equation}
Moreover, for any $(u, z) \in \sU \times \latentsymbolspace$,
\begin{equation}
    \frac{\indicate{u \in \sU}}{2^r}\indicate{F_q^{-1}(u) = z} = \frac{\indicate{u \in U(z)}}{2^r} = q(z \given x) \frac{\indicate{u \in U(z)}}{2^r q(z \given x)}.
\end{equation}
Thus,
\begin{align}
    Q(\{z_i\}_{i=1}^N, \{u_i\}_{i=1}^N, j \given x) &= \frac{\indicate{u_1 \in \sU}}{2^r} \prod_{i = 2}^N \indicate{T_i(u_1) = u_i}  \prod_{i = 1}^N  \indicate{F_q^{-1}(u_i) = z_i} \frac{w_j}{\sum_{i=1}^N w_i}\\
    &= \frac{\indicate{u_j \in \sU}}{2^r} \prod_{i \neq j} \indicate{T_i(T_j^{-1}(u_j)) = u_i}  \prod_{i = 1}^N  \indicate{F_q^{-1}(u_i) = z_i} \frac{w_j}{\sum_{i=1}^N w_i}\\
    &=  q(z_j \given x) \frac{\indicate{u_j \in U(z_j)}}{2^r q(z_j \given x)} \prod_{i \neq j} \indicate{T_i(T_j^{-1}(u_j)) = u_i}   \indicate{F_q^{-1}(u_i) = z_i} \frac{w_j}{\sum_{i=1}^N w_i}.
\end{align}
From this it directly follows that
\begin{equation}
    \label{eq:bbcis_estimator}
    \frac{P(x, \{z_i\}_{i=1}^N, \{u_i\}_{i=1}^N, j)}{Q(\{z_i\}_{i=1}^N, \{u_i\}_{i=1}^N, j \given x)} = \frac{1}{N} \frac{\sum_{i=1}^N w_i}{w_j} \frac{p(x, z_j)}{q(z_j\given x)} = \sum_{i=1}^N \frac{w_i}{N} = \sum_{i=1}^N\frac{1}{N}\frac{p(x,F_q^{-1}(T_i(u_1)))}{q(F_q^{-1}(T_i(u_1))\given x)} ,
\end{equation}
which is exactly the CIS estimator.

\begin{algorithm}[t]
\caption{Encode Procedure of BB-CIS}
\label{alg:bb_cis_encode}
\Procedure{\Encode{symbol $x$, message $m$}}{
    decode $u_1  $ with $  \uniform \{0, 1, \dots, 2^r-1\}$ \\
    assign $u_i = T_i(u)$ and  $z_i = F_q^{-1}(u_i)$ for $i \in \{1, \dots, N\}$ \\
    decode $j  $ with $  \cat\left(\tilde{w}_j\right)$ \\
    encode $u_j  $ with $  \uniform \{u : F_q^{-1}(u) = z_j\}$ \\
    encode $x$ with $  p(x \given z_j)$ \\
    encode $z_j  $ with $  p(z_j)$ \\
    encode $j  $ with $  \cat(1/N)$ \\
    \KwRet $m'$
}
\end{algorithm}
\paragraph{BB-CIS coder} Based on the extended latent space representation, the Bits-Back Coupled Importance Sampling (BB-CIS) coder is derived in Alg. \ref{alg:bb_cis_encode} and visualized in Fig. \ref{fig:bb_cis}. The expected net bit length for encoding a symbol $x$ is:
\begin{align}
 -\expect_{u_1, j}\left[\log \frac{P(x, \{z_i\}_{i=1}^N, \{u_i\}_{i=1}^N, j)}{Q(\{z_i\}_{i=1}^N, \{u_i\}_{i=1}^N, j \given x)}\right] = - \expect_{u_1}\left[ \log \left(\sum_{i=1}^N\frac{w_i}{N} \right) \right] =   - \expect_{u_1}\left[ \log \left(\sum_{i=1}^N\frac{1}{N}\frac{p(x,F_q^{-1}(T_i(u_1)))}{q(F_q^{-1}(T_i(u_1))\given x)} \right) \right].
\end{align}
The convergence of BB-CIS's net bitrate to $-\log p(x)$ will depend on the $T_i$. There are many possibilities, but one consideration is that both sender and receiver must share the $T_i$ (or a procedure for generating them). Another consideration is that the number of particles should never exceed $2^r$, $N \leq 2^r$. This is because we can execute numerical integration with a budget of $2^r$ particles. Assuming $T_i(u) = i$ for $i \in \intint{0}{2^r -1}$:
\begin{equation}
    \sum_{i=1}^N\frac{1}{N}\frac{p(x,F_q^{-1}(T_i(u_1)))}{q(F_q^{-1}(T_i(u_1))\given x)} = \sum_{u=0}^{2^r-1} \frac{1}{2^r}\frac{p(x,F_q^{-1}(u))}{q(F_q^{-1}(u)\given x)} = \sum_{z \in \latentsymbolspace} q(z\given x)\frac{p(x,z)}{q(z\given x)} = p(x).
\end{equation}
We briefly mention two possibilities:
\begin{enumerate}
    \item Let $T_i(u) = (u + k_i) \Mod{2^r}$ where $k_i \sim \uniform\intint{0}{2^r-1}$ i.i.d. generated by a pseudorandom number generator where the sender and receiver share the seed. This will enjoy a convergence rate similar to BB-IS, because $\{(u + k_i) \Mod{2^r}\}_{i=1}^N$ will be i.i.d. uniform on $\intint{0}{2^r - 1}$ for $u \sim \uniform\intint{0}{2^r-1}$.
    \item Inspired by the idea of numerical integration, let $T_i(u) = (u + k_i) \Mod{2^r}$ where $k_i$ is the $i$th element of a random permutation of $\intint{0}{2^r-1}$. With this scheme, BB-CIS is performing numerical integration on a random permutation of $\intint{0}{2^r-1}$. The rate at which the net bitrate converges to $-\log p(x)$ could clearly be made worse by inefficient permutations. Randomizing the order may help avoid such inefficient permutations. We did not experiment with this.
\end{enumerate}

\subsection{Bits-Back Annealed Importance Sampling}
\label{appendix:bb_ais}

Annealed importance sampling (AIS) generalizes importance sampling by introducing a path of intermediate distributions between the tractable base distribution $q(z \given x)$ and the unnormalized target distribution $p(x,z)$. For AIS with $N$ steps, for $i\in \{0,\dots, N\}$, define annealing distributions
\begin{align}
    & \pi_i (z) \propto f_i(z) = q(z \given x)^{1-\beta_i}p(x,z)^{\beta_i}\\
    & \beta_i \in [0,1], \beta_0=0, \beta_i < \beta_{i+1}, \beta_N=1.
\end{align}
Note that $\pi_0(z)=f_0(z)=q(z \given x)$. Then define the MCMC transition operator $\mathcal{T}_i$ that leave the intermediate distribution $\pi_i$ invariant and its reverse $\tilde{\mathcal{T}}_i$:
\begin{align}
    & \int \mathcal{T}_i (z' \given z) \pi_i (z) \,dz=\pi_i (z') \\
    & \tilde{\mathcal{T}}_i(z' \given z)=\mathcal{T}_i (z \given z') \frac{\pi_i(z')}{\pi_i (z)}=\mathcal{T}_i (z \given z') \frac{f_i(z')}{f_i (z)} \label{eq:ais_reverse_kernel}.
\end{align}
Note, $\tilde{\mathcal{T}}_i$ is a normalized distribution. AIS samples a sequence of latents $\{z_i\}_{i=1}^N$ from base distribution and the MCMC transition kernels (see the Q process in Alg. \ref{alg:ais_extend_rep}) and obtains an unbiased estimate of $p(x)$ as the importance weight over the extended space.
\begin{equation}
    \hat{p}_N(x)=\frac{f_1(z_1)f_2(z_2)\dots f_N(z_N)}{f_0(z_1)f_1(z_2)\dots f_{N-1}(z_N)}.
\end{equation}
The corresponding variational bound is the log importance weight:
\begin{equation}
    - \expect_{\{z_i\}_{i=1}^N} \left[ \log \frac{f_1(z_1)f_2(z_2)\dots f_N(z_N)}{f_0(z_1)f_1(z_2)\dots f_{N-1}(z_N)} \right] \geq -\log p(x).
\end{equation}

\begin{algorithm}[t]
\caption{Extended Latent Space Representation of Annealed Importance Sampling}
\label{alg:ais_extend_rep}

\begin{minipage}[t]{0.46\textwidth}
\Process{$Q(\mathcal{Z}  \given  x)$}{
    sample $z_1 \sim q(z_1 \given x) = \pi_0 (z_1)$ \\
    \For{$i = 1,\dots,N-1 $}{
    sample $z_{i+1} \sim \mathcal{T}_i(z_{i+1}  \given  z_i)$}
    \KwRet $\{z_i\}_{i=1}^N$
}
\end{minipage}
\begin{minipage}[t]{0.46\textwidth}
\Process{$P(x, \mathcal{Z})$}{
    sample $z_N \sim p(z_N)$ \\
    \For{$i = N-1,\dots,1$}{sample $z_{i} \sim \tilde{\mathcal{T}}_{i}(z_i  \given  z_{i+1})$}
    sample $x \sim p(x  \given  z_N )$ \\
    \KwRet $x, \{z_i\}_{i=1}^N$
}
\hfill
\end{minipage}
\end{algorithm}
\paragraph{Extended latent space representation}
The extended space representation of AIS is derived in \citep{neal2001annealed}, as presented in Alg.  \ref{alg:ais_extend_rep}. The extended latent variables contain all the latents $\{z_i\}_{i=1}^N$. Briefly, given $x$, the distribution $Q$ first samples $z_1$ from the base distribution $q(z_1 \given x)$ and then sample $\{z_i\}_{i=2}^N$ sequentially from the transition kernel $\mathcal{T}_i (z_{i+1} \given z_i)$. The distribution $P$ first samples $z_N$ from the prior $p(z_N)$, then samples $\{z_i\}_{i=N-1}^1$ from the reverse transition kernel $\tilde{\mathcal{T}}_{i}(z_i  \given  z_{i+1})$ in the reverse order, and finally samples $x$ from $p(x \given z_N)$. The proposal distribution and the target distribution can be derived as followed:
\begin{align}
    Q(\{z_i\}_{i=1}^N)&=q (z_1  \given  x) \mathcal{T}_1(z_2  \given  z_1) \mathcal{T}_2(z_3  \given  z_2) \dots \mathcal{T}_{N-1}(z_N  \given  z_{N-1}) \\
    &= f_0 (z_1) \mathcal{T}_1(z_2  \given  z_1) \mathcal{T}_2(z_3  \given  z_2) \dots \mathcal{T}_{N-1}(z_N  \given  z_{N-1})
\end{align}
\begin{align}
    P(x, \{z_i\}_{i=1}^N)&=p(x, z_N) \tilde{\mathcal{T}}_{N-1}(z_{N-1}  \given  z_{N}) \tilde{\mathcal{T}}_{N-2}(z_{N-2}  \given  z_{N-1}) \dots \tilde{\mathcal{T}}_{1}(z_{1}  \given  z_{2})\\
    &=f_N(z_N) \tilde{\mathcal{T}}_{N-1}(z_{N-1}  \given  z_{N}) \tilde{\mathcal{T}}_{N-2}(z_{N-2}  \given  z_{N-1}) \dots \tilde{\mathcal{T}}_{1}(z_{1}  \given  z_{2}).
\end{align}
Now note
\begin{align}
     \frac{P(x, \{z_i\}_{i=1}^N)}{Q(\{z_i\}_{i=1}^N)}&= \frac{f_N(z_N)\tilde{\mathcal{T}}_{N-1}(z_{N-1} | z_{N}) \tilde{\mathcal{T}}_{N-2}(z_{N-2} | z_{N-1}) \dots \tilde{\mathcal{T}}_{1}(z_{1} | z_{2})}{f_0 (z_1) \mathcal{T}_1(z_2 | z_1) \mathcal{T}_2(z_3 | z_2) \dots \mathcal{T}_{N-1}(z_N | z_{N-1})} \\
     &=\frac{f_1(z_1)f_2(z_2)\dots f_N(z_N)}{f_0(z_1)f_1(z_2)\dots f_{N-1}(z_N)},
\end{align}
which exactly gives us the AIS estimator. Note that we have used \eqref{eq:ais_reverse_kernel} above.
\begin{algorithm}[t]
\caption{Encode Procedure of BB-AIS}
\label{alg:bb_ais_encode}
\Procedure{\Encode{symbol $x$, message $m$}}{
    decode $z_1 $ with $ q(z_1 \given x) = \pi_0 (z_1)$ \\
    \For{$i = 1,\dots,N-1 $}{
    decode $z_{i+1} $ with $ \mathcal{T}_i(z_{i+1}  \given  z_i)$}
    encode $x $ with $ p(x  \given  z_N )$ \\
    \For{$i = 1,\dots, N-1$}{
    encode $z_{i} $ with $ \tilde{\mathcal{T}}_{i}(z_i  \given  z_{i+1})$}
    encode $z_N $ with $ p(z_N)$ \\
    \KwRet $m'$
}
\end{algorithm} \begin{algorithm}[t]
\caption{Encode Procedure of BB-AIS with BitSwap}
\label{alg:bb_ais_bitswap_encode}
\Procedure{\Encode{symbol $x$, message $m$}}{
    decode $z_1 $ with $ q(z_1 \given x) = \pi_0 (z_1)$ \\
    \For{$i = 1,\dots,N-1 $}{
    decode $z_{i+1} $ with $ \mathcal{T}_i(z_{i+1}  \given  z_i)$\\
    encode $z_{i} $ with $ \tilde{\mathcal{T}}_{i}(z_i  \given  z_{i+1})$}
    encode $x $ with $ p(x  \given  z_N )$ \\
    encode $z_N $ with $ p(z_N)$ \\
    \KwRet $m'$
}
\end{algorithm}
\begin{figure*}[t]
    \centering
    \begin{subfigure}[t]{0.55\textwidth}
        \includegraphics{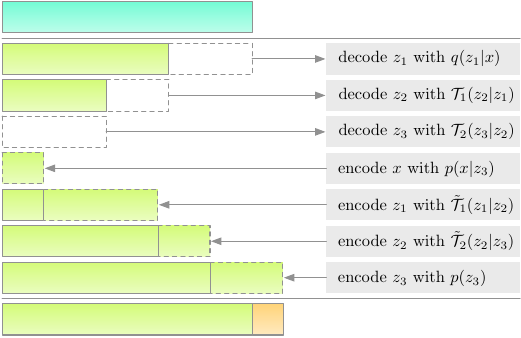}
        \caption{Bits-Back Annealed Importance Sampling}
        \label{fig:bbais}
    \end{subfigure}
    \hfill
    \begin{subfigure}[t]{0.44\textwidth}
        \includegraphics{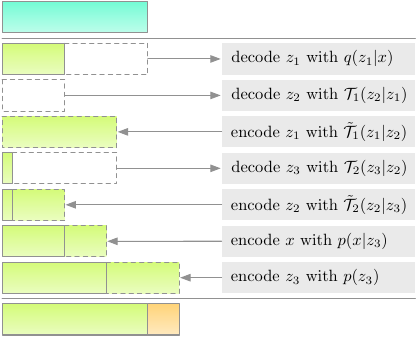}
        \caption{Bits-Back Annealed Importance Sampling with BitSwap}
        \label{fig:bbais_bitswap}
    \end{subfigure}
    \caption{The encoding schemes of BB-AIS and BB-AIS with BitSwap both with $N=3$ steps. Encoding a symbol $x$ with BB-AIS incurs a very large initial bit cost that scales like $\gO(N)$. This can be significantly reduced by applying the BitSwap trick. }
\end{figure*}

\paragraph{BB-AIS coder}
Based on the extended space representation of AIS, the Bits-Back Annealed Importance Sampling coder is derived in Alg. \ref{alg:bb_ais_encode} and visualized in Fig. \ref{fig:bbais}. The expected net bit length for encoding a symbol $x$ is:
\begin{align}
 -\expect_{\{z_i\}_{i=1}^N}\left[ \frac{\log P(x, \{z_i\}_{i=1}^N)}{\log Q(\{z_i\}_{i=1}^N \given x)}  \right] =   - \expect_{\{z_i\}_{i=1}^N} \left[ \log \frac{f_1(z_1)f_2(z_2)\dots f_N(z_N)}{f_0(z_1)f_1(z_2)\dots f_{N-1}(z_N)} \right].
\end{align}
Ignoring the dirty bits issue, i.e., $\{z_i\}_{i=1}^N \sim Q(\{z_i\}_{i=1}^N \given x)$, the expected net bit length exactly achieves the negative AIS bound.

The BB-AIS coder has several practical issues. First is the computational cost. Since the intermediate distributions are usually not factorized over the latent dimensions, it is very inefficient to encode and decode the latents with entropy coders for high dimensional latents. Second is the increased initial bit cost as with BB-IS. More precisely, the initial bits that BB-AIS requires is as followed:
\begin{equation}
    - \log q (z_1 \given x) - \sum_{i=1}^{N-1}\log \mathcal{T}_i(z_{i+1}  \given  z_i),
\end{equation}
which scales like $\gO (N)$. However, because the structure of BB-AIS is very similar to the hierarchical latent variable modelss with Markov structure in \citep{kingma2019bit}, the BitSwap trick can be applied with BB-AIS to reduce the initial bit cost. The BB-AIS coder with BitSwap is derived in Alg. \ref{alg:bb_ais_bitswap_encode} and visualized in \ref{fig:bbais_bitswap}. In particular, note that the latents $\{z_i\}_{i=1}^N$ in BB-AIS are encoded and decoded both in the forward time order, we can interleave the encode/decode operations such that $z_i$ are encoded with $z_{i+1}$ as soon as $z_{i+1}$ are decoded. Thus there are always at least $-\log \tilde{\mathcal{T}}_{i}(z_i  \given  z_{i+1})$ available for decoding $z_{i+2}$ at the next step, and the initial bit cost is bounded by:
\begin{equation}
    - \log q (z_1 \given x) - \log \mathcal{T}_1(z_{2}  \given  z_1) + \sum_{i=1}^{N-2}\max (0, - \log \mathcal{T}_{i+1}(z_{i+2}  \given  z_{i+1}) + \log \tilde{\mathcal{T}}_{i}(z_i  \given  z_{i+1}) ).
\end{equation}
Although BitSwap helps to reduce the initial bit cost, we find that it suffers more from the dirty bits issue than the naive implementation and affects the expected net bit length (see Fig. \ref{fig:cleanliness_plot_toy_mixture_only_ais} for an empirical study). Therefore, using BB-AIS with BitSwap leads to a trade-off between net bitrate distortion and inital bit cost that depends on the number of symbols to be encoded.

\subsection{Bits-Back Sequential Monte Carlo}
\newcommand{\smcancestors}[1]{\boldsymbol{A}_{#1}}
\newcommand{\smcstates}[1]{\boldsymbol{Z}_{#1}}
\newcommand{\smcobs}[1]{\boldsymbol{x}_{#1}}
\newcommand{\smclatent}[1]{\boldsymbol{z}_{#1}}
\newcommand{\smcancestrajec}[2]{\boldsymbol{\tau}_{#1}^{#2}}

Sequential Monte Carlo (SMC) is a particle filtering method that combines importance sampling with resampling, and its estimate of marginal likelihood typically converges faster than importance sampling for sequential latent variable models \cite{cerou2011nonasymptotic, berard2014lognormal}. Suppose the observations are a sequence of $T$ random variables $\smcobs{T} \in \symbolspace^T$, where $\smcobs{t} := x_{1:t}$. Sequential latent variable models introduce a sequence of $T$ (unobserved) latent variables $\smclatent{T} \in \latentsymbolspace^T$ associated with $\smcobs{T}$, where $\smclatent{t} := z_{1:t}$. We assume the joint distribution $p(\smcobs{T}, \smclatent{T})$ can be factored as:
\begin{equation}
    p(\smcobs{T}, \smclatent{T}) = \prod_{t=1}^{T}f(z_t \given \smcobs{t-1}, \smclatent{t-1})g(x_t \given \smcobs{t-1}, \smclatent{t}),
\end{equation}
where $f$ and $g$ are (generalized) transition and emission distributions, respectively. For the case $t=1$, $f(z_1 \given \emptyset, \emptyset)=\mu(z_1)$ reduces to the prior distribution and $g(x_1 \given \emptyset, z_1)$ only conditions on $z_1$. We also assume the approximate posterior $q(\smclatent{T} \given \smcobs{T})$ can be factorized as:
\begin{equation}
    q(\smclatent{T} \given \smcobs{T})=\prod_{t=1}^{T} q(z_t \given \smcobs{t}, \smclatent{t-1}).
\end{equation}
For the case $t=1$, $q(z_1 \given x_1, \emptyset)$ only conditions on $x_1$.

SMC maintains a population of particle states $\smcstates{t}:=z_{1:t}^{1:N}$. Here we use subscript for indexing timestep $t$ and superscript for indexing particle $i$. In this appendix we describe the version of SMC that resamples at every iteration. Adaptive resampling schemes are possible \citep{doucet2001introduction}, but we omit them for clarity.

At each time step $t$, each particle independently samples an extension $z_t^i \sim q(z_t^i \given \smcobs{t}, \smcancestrajec{t-1}{i})$ where $\smcancestrajec{t-1}{i}$ is the ancestral trajectory of the $i$th particle at timestep $t$ (will be described in details later). The sampled extension $z_t^i$ is appended to $\smcancestrajec{t-1}{i}$ to form the trajectory of $i$th particle as $(\smcancestrajec{t-1}{i}, z_t^i)$. Then the whole population is resampled with probabilities in proportion to importance weights defined as followed:
\begin{align}
    &w_t^i = \frac{f(z_t^i \given \smcobs{t-1}, \smcancestrajec{t-1}{i})g(x_t \given \smcobs{t-1}, (\smcancestrajec{t-1}{i}, z_t^i))}{q(z_t^i \given \smcobs{t}, \smcancestrajec{t-1}{i})} \\
    &\tilde{w}_t^i = \frac{w_t^i}{\sum_{i=1}^N w_t^i}.
\end{align}
In particular, each particle samples a `parent' index $A_{t}^i \sim \cat(\tilde{w}_t^i)$. At the next timestep $t+1$, the $i$th particle `inherits' the ancestral trajectory of the $A_{t}^i$th particle. More precisely,
\begin{equation}
    \smcancestrajec{t}{i} = (\smcancestrajec{t-1}{A_t^i}, z_t^{A_{t}^i}). \label{eq:smc_recur}
\end{equation}
Thus, at timestep $t+1$, given the parent index $A_{t}^i$, as well as the collection of particle states $\smcstates{t}$ and previous ancestral indices $\smcancestors{t-1} := A_{1:t-1}^{1:N}$, one can trace back the whole ancestral trajectory $\smcancestrajec{t}{i}$ by recursively applying \eqref{eq:smc_recur}.

\begin{figure}[t]
    \centering
    \includegraphics{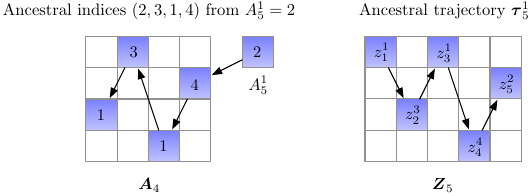}
    \caption{Trace back the ancestral trajectory $\boldsymbol{\tau}_{t-1}^i$ for the particle $i=1$ at timestep $t=6$.}
    \label{fig:trace_back}
\end{figure}

For brevity and consistency, we provide the pseudocode (Alg. \ref{alg:smc_trace_back}) and visualization (Fig. \ref{fig:trace_back}) for tracing back the ancestral trajectory $\smcancestrajec{t-1}{i}$ for the $i$th particle at timestep $t$. In particular, we first obtain the ancestral index $B_k^i$ at each previous timestep $k$ by using the fact
\begin{equation}
    B_{k}^i = A_{k}^{B_{k+1}^i} \label{eq:smc_ances_indices}.
\end{equation}
Thus we can obtain a series of ancestral indices $(B_{1}^i, \dots, B_{t-1}^i)$ in the backward time order and then obtain the ancestral trajectory by indexing $\smcstates{t-1}$. Note that Alg. \ref{alg:smc_trace_back} is only for helping understand the SMC algorithm, in practice, typically the ancestral trajectories are book-kept and updated along the way using \eqref{eq:smc_ances_indices}.

\begin{algorithm}[t]
\caption{Tracing Back the Ancestral Trajectory $\smcancestrajec{t-1}{i}$ for the $i$th Particle at Timestep $t$}
\label{alg:smc_trace_back}
\SetKwFunction{Trace}{TraceBack}{}{}

\Procedure{\Trace{$\smcstates{t-1}, \smcancestors{t-2}$, $A_{t-1}^i$}}{
    assign $B_{t-1}^i = A_{t-1}^i$ \\
    \For{$k=t-2, \dots, 1$}{
        assign $B_{k}^i = A_{k}^{B_{k+1}^i}$
    }
    assign $\smcancestrajec{t-1}{i}=(z_1^{B_1^i}, z_2^{B_2^i}, \dots, z_{t-1}^{B_{t-1}^i})$ \\
    \KwRet $\smcancestrajec{t-1}{i}$
}
\end{algorithm}

SMC obtains an unbiased estimate of $p(\smcobs{T})$ using the intermediate importance weights \citep[Proposition 7.4.1]{delmoral2004}:
\begin{equation}
    \hat{p}_N(\smcobs{T})=\prod_{t=1}^T \left(\frac{1}{N}\sum_{i=1}^N w_t^i \right).
\end{equation}
The corresponding variational bound \citep[FIVO,][]{maddison2017filtering, naesseth2018vsmc, le2018auto} is
\begin{equation}
    -\expect_{\smcstates{T}, \smcancestors{T-1}} \left [ \sum_{t=1}^T \log \left(\frac{1}{N}\sum_{i=1}^N w_t^i \right) \right] \geq - \log p(\smcobs{T}).
\end{equation}

\begin{algorithm}[t]
\caption{Extended Latent Space Representation of Sequential Monte Carlo}
\label{alg:smc_extend_rep}
\Process{$Q(\mathcal{Z} \given \smcobs{T})$}{
    \For{$t = 1, \ldots, T$}{
        \If{$t \neq 1$}{
            \For{$i = 1, \ldots, N$}{
            sample $A^i_{t-1} \sim \cat(\tilde{w}_{t-1})$
            }
            assign $\smcancestors{t-1} = \left[\smcancestors{t-2}, A_{t-1}^{1:N} \right]$
        }
        \For{$i = 1, \ldots, N$}{
            assign $\smcancestrajec{t-1}{i} = \mathtt{TraceBack}(\smcstates{t-1}, \smcancestors{t-2}, A_{t-1}^i)$\\
            sample $z_t^i \sim q(z_t^i  \given  \smcobs{t}, \smcancestrajec{t-1}{i})$
        }
        assign $\smcstates{t} = \left[\smcstates{t-1}, z_{t}^{1:N}\right]$
        }
    sample $j \sim \cat(\tilde{w}_T)$ \\
    \KwRet $\smcstates{T}, \smcancestors{T-1}, j$
}
~\\
\Process{$P(\smcobs{T}, \mathcal{Z})$}{
    \For{$t = 1, \ldots, T$}{
        sample $B_t \sim \cat(1/N)$ \\
        \If{$t \neq 1$}{
            assign $A_{t-1}^{B_t} = B_{t-1}$ \\
            \For{$i \neq B_t$}{
                sample $A^i_{t-1} \sim \cat(\tilde{w}_{t-1})$ \\
            }
            assign $\smcancestors{t-1} = \left[\smcancestors{t-2}, A_{t-1}^{1:N} \right]$
        }
        assign $\smcancestrajec{t-1}{B_t} = \mathtt{TraceBack}(\smcstates{t-1}, \smcancestors{t-2}, B_{t-1})$\\
        sample $z_t^{B_t} \sim  f(z_t^{B_t} \given \smcobs{t-1}, \smcancestrajec{t-1}{B_t})$ \\
        sample $x_t \sim  g(x_t \given \smcobs{t-1}, (\smcancestrajec{t-1}{B_t}, z_t^{B_t}))$ \\
        \For{$i \neq B_t$}{
            assign $\smcancestrajec{t-1}{i} = \mathtt{TraceBack}(\smcstates{t-1}, \smcancestors{t-2}, A^i_{t-1})$\\
            sample $z_t^i \sim q(z_t^i  \given  \smcobs{t}, \smcancestrajec{t-1}{i})$
        }
        assign $\smcstates{t} = \left[\smcstates{t-1}, z_{t}^{1:N}\right]$ }
    assign $j = B_T$ \\
    \KwRet $\smcobs{T}, \smcstates{T}, \smcancestors{T-1}, j$
}
\end{algorithm}

\paragraph{Extended latent space representation}
The extended space representation of SMC is derived in \citep{andrieu2010particle}, as presented in Alg. \ref{alg:smc_extend_rep}. The extended latent space variables contain the particle states $\smcstates{T}$, ancestral indices $\smcancestors{T-1}$, and a particle index $j$ used to pick one special particle trajectory. Briefly, the $Q$ distribution samples $\smcstates{T}$ and $\smcancestors{T-1}$ in the same manner as SMC, with an additional step that samples the particle index $j$ with probability proportional to the importance weight $w_T$. The $P$ distribution selects the ancestral indices of the special particle $(B_1, \dots, B_T)$ uniformly, and samples the special particle trajectory $\smclatent{T}^*=(z_1^{B_1}, \dots, z_T^{B_T})$ and the observations $\smcobs{T}$ jointly with the underlying model distribution. The remaining particle states and ancestral indices are sampled with the same distribution as $Q$. Note that the special particle trajectory $\smclatent{T}^*=\mathtt{TraceBack}(\smcstates{T}, \smcancestors{T-1}, j)$.

From Alg. \ref{alg:smc_extend_rep}, the proposal distribution and the target distribution and be derived as followed:
\begin{align}
    Q(\smcstates{T}, \smcancestors{T-1}, j \given \smcobs{T})=\tilde{w}_T^j \prod_{i=1}^N q(z_1^i|x_1) \prod_{t=2}^T \prod_{i=1}^N \tilde{w}_{t-1}^{A^i_{t-1}} q(z_t^i  \given  \smcobs{t}, \smcancestrajec{t-1}{i})
\end{align}
\begin{align}
    P(\smcobs{T}, \smcstates{T}, \smcancestors{T-1}, j) &=\frac{1}{N^T}\mu(z_1^{B_1}) g(x_1|z_1^{B_1}) \prod_{t=2}^T f(z_t^{B_t} \given \smcobs{t-1}, \smcancestrajec{t-1}{B_t}) g(x_t \given \smcobs{t-1}, (\smcancestrajec{t-1}{B_t}, z_t^{B_t}))\nonumber\\
    &~~~~\prod_{i \neq B_1} q(z_1^i|x_1) \prod_{t=2}^T\prod_{i \neq B_t} \tilde{w}_{t-1}^{A^i_{t-1}} q(z_t^i  \given  \smcobs{t}, \smcancestrajec{t-1}{i})\\
    &=\frac{1}{N^T}p(\smcobs{T}, \smclatent{T}^*)\prod_{i \neq B_1} q(z_1^i|x_1) \prod_{t=2}^T\prod_{i \neq B_t} \tilde{w}_{t-1}^{A^i_{t-1}} q(z_t^i  \given  \smcobs{t}, \smcancestrajec{t-1}{i}).
\end{align}

Now note,
\begin{align}
    \frac{P(\smcobs{T}, \smcstates{T}, \smcancestors{T-1}, j)}{Q(\smcstates{T}, \smcancestors{T-1}, j \given \smcobs{T})}&=\frac{1}{N^T \tilde{w}_T^j}\frac{p(\smcobs{T}, \smclatent{T}^*)}{q(z_1^{B_1}|x_1) \prod_{t=2}^T \tilde{w}_{t-1}^{A^{B_t}_{t-1}} q(z_t^{B_t}  \given  \smcobs{t}, \smcancestrajec{t-1}{B_t})} \\
    &=\frac{1}{N^T \tilde{w}_T^j}\frac{p(\smcobs{T}, \smclatent{T}^*)}{q(z_1^{B_1}|x_1) \prod_{t=2}^T \tilde{w}_{t-1}^{B_{t-1}} q(z_t^{B_t}  \given  \smcobs{t}, \smcancestrajec{t-1}{B_t})} \\
    &=\frac{1}{N^T\prod_{t=1}^T \tilde{w}_{t}^{B_{t}}} \frac{\mu(z_1^{B_1}) g(x_1|z_1^{B_1}) \prod_{t=2}^T f(z_t^{B_t} \given \smcobs{t-1}, \smcancestrajec{t-1}{B_t}) g(x_t \given \smcobs{t-1}, (\smcancestrajec{t-1}{B_t}, z_t^{B_t}))}{q(z_1^{B_1}|x_1) \prod_{t=2}^T q(z_t^{B_t}  \given  \smcobs{t}, \smcancestrajec{t-1}{B_t})} \\
    &=\frac{\prod_{t=1}^T w_{t}^{B_{t}}}{N^T\prod_{t=1}^T \tilde{w}_{t}^{B_{t}}} \\
    &=\prod_{t=1}^T(\frac{1}{N}\sum_{i=1}^N w_t^i),
\end{align}
which exactly gives us the SMC estimator. Note that we have used \eqref{eq:smc_ances_indices} and $B_T=j$ in the above derivation.

\begin{algorithm}[t]
\caption{Encode Procedure of BB-SMC}
\label{alg:bb_smc_encode}
\Procedure{\Encode{symbol $\smcobs{T}$, message $m$}}{
    \For{$t = 1, \ldots, T$}{
        \If{$t \neq 1$}{
            \For{$i = 1, \ldots, N$}{
            decode $A^i_{t-1}  $ with $  \cat(\tilde{w}_{t-1})$
            }
            assign $\smcancestors{t-1} = \left[\smcancestors{t-2}, A_{t-1}^{1:N} \right]$
        }
        \For{$i = 1, \ldots, N$}{
            assign $\smcancestrajec{t-1}{i} = \mathtt{TraceBack}(\smcstates{t-1}, \smcancestors{t-2}, A_{t-1}^i)$\\
            decode $z_t^i  $ with $  q(z_t^i  \given  \smcobs{t}, \smcancestrajec{t-1}{i})$
        }
        assign $\smcstates{t} = \left[\smcstates{t-1}, z_{t}^{1:N}\right]$
        }
    decode $j  $ with $  \cat(\tilde{w}_T)$ \\
    set the ancestral lineage $B_T = j$ and $B_t = A_t^{B_{t+1}}$ for $t=T-1, \dots, 1$ \\
    \For{$t = T, \ldots, 1$}{
        \For{$i \neq B_t$}{
            encode $z_t^i  $ with $  q(z_t^i  \given  \smcobs{t}, \smcancestrajec{t-1}{i})$
        }
        encode $x_t  $ with $   g(x_t \given \smcobs{t-1}, (\smcancestrajec{t-1}{B_t}, z_t^{B_t}))$ \\
        encode $z_t^{B_t}  $ with $   f(z_t^{B_t} \given \smcobs{t-1}, \smcancestrajec{t-1}{B_t})$ \\
        \If{$t \neq 1$}{
            \For{$i \neq B_t$}{
                encode $A^i_{t-1}  $ with $  \cat(\tilde{w}_{t-1})$ \\
            }
        }
        encode $B_t  $ with $  \cat(1/N)$ \\
         }
    \KwRet $m'$
}

\end{algorithm}

\begin{figure}
    \centering
    \includegraphics[width=.95\linewidth]{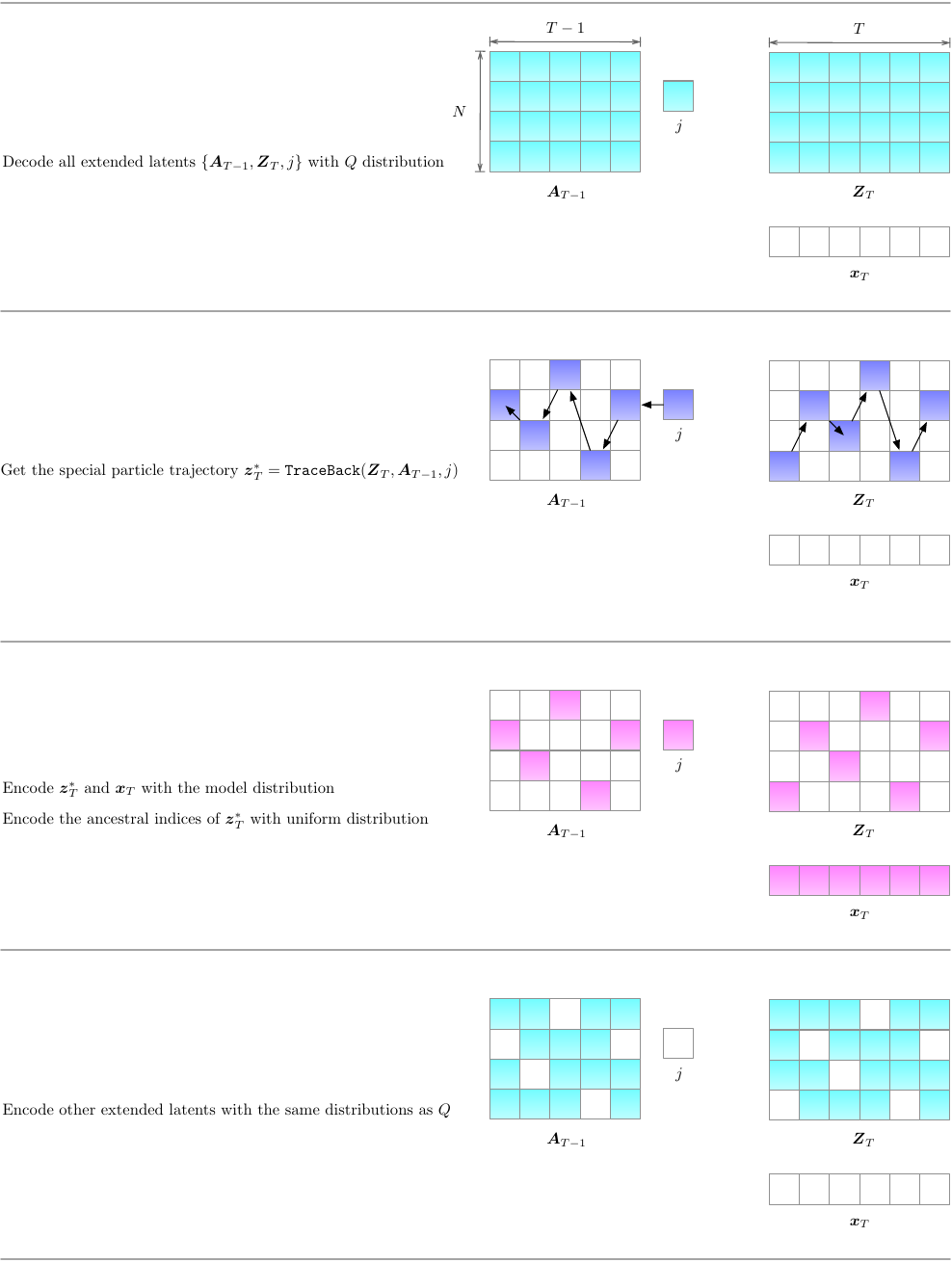}
    \caption{The visualization of the encode procedure of BB-SMC.}
    \label{fig:bb_smc}
\end{figure}

\paragraph{BB-SMC coder}
Based on the extended space representation of SMC, the Bits-Back Sequential Monte Carlo coder is derived in Alg. \ref{alg:bb_smc_encode} and visualized in Fig. \ref{fig:bb_smc}.
Intuitively, BB-SMC first decodes all the extended latent variables $\mathcal{Z}=\{\smcstates{T}, \smcancestors{T-1}, j\}$ in the forward time order. Then it picks a special particle and traces its trajectory $\smclatent{T}^*$ backward in time. The distribution of the special trajectory can be seen as a non-parametric approximation of the true posterior. The special trajectory $\smclatent{T}^*$ is used to encode the sequential observations $\smcobs{T}$ and is itself encoded with with the model distribution. The special trajectory's ancestral indices are encoded using a uniform distribution. All other extended latent variables are encoded back with the same distribution as decoding and in the backward time order. The expected net bit length for encoding a symbol $\smcobs{T}$ is:
\begin{align}
 -\expect_{\smcstates{T}, \smcancestors{T-1}, j}\left[\frac{\log P(\smcobs{T}, \smcstates{T}, \smcancestors{T-1}, j)}{\log Q(\smcstates{T}, \smcancestors{T-1}, j \given \smcobs{T})}  \right] =   -\expect_{\smcstates{T}, \smcancestors{T-1}} \left [ \sum_{t=1}^T \log \left(\frac{1}{N}\sum_{i=1}^N w_t^i \right) \right].
\end{align}
Ignoring the dirty bits issue, the expected net bit length exactly achieves the negative FIVO bound.

Note that our BB-SMC scheme can be easily extended to adaptive resampling setting where the particles are only resampled when a certain resampling criteria is satisfied. In adaptive schemes, the importance weights accumulate multiplicatively over time (between resampling events). One typical adaptive resampling criteria is applying resampling if the effective sample size (ESS) of the particles drops below $N/2$. Details can be found in \citep{doucet2001introduction}.

Adaptive resampling is possible, because the same resampling decisions made at encode time by the sender can be exactly recovered by the receiver. The encoding process of the sender produces the SMC state in the forward direction of time. The receiver also reconstructs the the SMC state in the forward direction of time. Thus, if the receiver has the same resampling criteria as the sender, they can recover exactly the SMC forward pass of the sender. Thus, we can make two major modifications to incorporate adaptive resampling to BB-SMC. First is that the parent indices $A_{t-1}^i$ are only decoded (and then encoded back) when the resampling criteria is satisfied at each timestep $t$, otherwise each particle inherits itself, i.e., set $A_{t-1}^i=i$. Second is that when resampling is not performed, the importance weights need to be accumulated and used for resampling next time. After resampling, the accumulated importance weights are reset to uniform for all particles.

\paragraph{BB-CSMC coder}
\begin{algorithm}[t]
\caption{Encode Procedure of BB-CSMC}
\label{alg:bb_csmc_encode}
\Procedure{\Encode{symbol $\smcobs{T}$, message $m$}}{
    \For{$t = 1, \ldots, T$}{
        \If{$t \neq 1$}{
            decode $v_{t-1}$ with $\uniform \{0, \cdots, 2^r - 1 \}$ \\
            assign $p_{t-1} = \cat(\tilde{w}_{t-1})$\\
            \For{$i = 1, \ldots, N$}{
            assign $v_{t-1}^i=R_{t-1}^i (v_{t-1})$ \\
            assign $A_{t-1}^i=F^{-1}_{p_{t-1}} (v_{t-1}^i)$
            }
            assign $\smcancestors{t-1} = \left[\smcancestors{t-2}, A_{t-1}^{1:N} \right]$
        }
        decode $u_t$ with $\uniform \{0, \cdots, 2^r - 1 \}$ \\
        \For{$i = 1, \ldots, N$}{
            assign $\smcancestrajec{t-1}{i} = \mathtt{TraceBack}(\smcstates{t-1}, \smcancestors{t-2}, A_{t-1}^i)$\\
            assign $u_t^i=T_t^i (u_t)$, $q_t^i = q(z_t^i  \given  \smcobs{t}, \smcancestrajec{t-1}{i})$ \\
            assign $z_t^i=F^{-1}_{q_t^i} (u_t^i)$
        }
        assign $\smcstates{t} = \left[\smcstates{t-1}, z_{t}^{1:N}\right]$
        }
    decode $j  $ with $  \cat(\tilde{w}_T)$ \\
    set the ancestral lineage $B_T = j$ and $B_t = A_t^{B_{t+1}}$ for $t=T-1, \dots, 1$ \\
    \For{$t = T, \ldots, 1$}{
        encode $u_t^{B_t}$ with $\uniform \{u: F^{-1}_{q_t^{B_t}} (u) = z_t^{B_t} \}$ \\
        encode $x_t  $ with $   g(x_t \given \smcobs{t-1}, (\smcancestrajec{t-1}{B_t}, z_t^{B_t}))$ \\
        encode $z_t^{B_t}  $ with $   f(z_t^{B_t} \given \smcobs{t-1}, \smcancestrajec{t-1}{B_t})$ \\
        \If{$t \neq 1$}{
            encode $v^{B_t}_{t-1}  $ with $\uniform \{v: F^{-1}_{p_{t-1}} (v) = A_{t-1}^{B_t} \}$
        }
        encode $B_t  $ with $  \cat(1/N)$ \\
         }
    \KwRet $m'$
}

\end{algorithm}

As with BB-IS, BB-SMC also suffers from a increased initial bits cost equal to $-\log Q(\smcstates{T}, \smcancestors{T-1}, j)$ that scales like $\gO(NT)$, in contrast to the $\gO (T)$ initial bit cost of BB-ELBO. Similarly, we can derive a coupled variant of BB-SMC which is called Bits-Back Coupled Sequential Monte Carlo (BB-CSMC).

BB-CSMC reparameterizes the particle states $z_t^i$ as deterministic functions of uniform random variables $u_t^i$ which are coupled by a common uniform $u_t$. At each timestep $t$, instead of directly decoding $z_t^i$ with their approximate posterior $q_t^i$, BB-CSMC first decodes $u_t$ and then obtain $u_t^i$ with bijective functions $T_t^i$, i.e., $u_t^i=T_t^i(u_t)$. Then the particle states $z_t^i$ are obtained as $z_t^i=F^{-1}_{q_t^i} (u_t^i)$, where the functions $F^{-1}_{q_t^i}$ are defined as \eqref{eq:gen_inv_cdf}. This is not enough to sufficiently reduce the $\gO(NT)$ initial bit cost since the parent indices $A_{t-1}^i$ are also decoded (and thus require some initial bits) for each particle. Therefore, similarly for $A_{t-1}^i$, we also need to introduce the common uniform $v_{t-1}$ and the bijective functions $R_{t-1}^i$ to obtain $v_{t-1}^i=R_{t-1}^i (v_{t-1})$. $A_{t-1}^i$ are obtained as $A_{t-1}^i=F^{-1}_{p_{t-1}} (v_{t-1}^i)$ where $p_{t-1}$ is the categorical distribution defined by normalized importance weights $\tilde{w}_{t-1}$. This reduces the initial bit cost to $(2T-1)r-\log(\tilde{w}_T)$. Note that $r$ is lower bounded by $\log N$ because $2^r$ should be larger than $N$, the initial bit cost roughly scales like $\gO(T\log N)$.

The encoding process of BB-CSMC should also be calibrated to match the modified decoding process, as with BB-CIS. In particular, after decoding the special particle index $j$, BB-CSMC only encodes $(x_t, B_{t}, v_{t-1}^{B_t}, u_t^{B_t}, z_t^{B_t})$ associated with the special particle trajectory. The uniform $v_{t-1}^{B_t}$ and $u_t^{B_t}$ are encoded using uniform distributions over restricted sets mapped from $A_{t-1}^{B_t}=B_{t-1}$ and $z_t^{B_t}$, respectively.

One can easily compute the net bitrate of BB-CSMC is:
\begin{equation}
-\expect_{\{u_t\}_{t=1}^T, \{v_t\}_{t=1}^{T-1}} \left [ \sum_{t=1}^T \log \left(\frac{1}{N}\sum_{i=1}^N w_t^i \right) \right]
\end{equation}
which is comparable to BB-SMC but uses only $\gO (T\log N)$ initial bits.

 \section{Computational Cost}
\label{appendix:compcost}

The computational requirements of our McBits coders can be distilled into two components: calculation of the distributions used (which usually involves neural network computation) and the actual encoding/decoding operations themselves. In our implementations of McBits the neural network computations tend to dominate.

For most McBits coders (e.g., BB-IS and BB-AIS), the computational complexity scales like $\gO(N)$, where $N$ is the number of particles or AIS steps. In BB-IS for example, the neural network for the approximate posterior needs to be computed once for each encoded symbol, while the neural net for the conditional likelihood is executed $N$ times, once for each particle, in order to compute the weights for the categorical sampler.

In BB-SMC, we are required to decode/encode $N$ times from a categorical distribution of alphabet size $N$ at each timestep. Naively implemented, this has computational complexity $\gO(TN^2)$ (in contrast to the $\gO(T)$ computational complexity of BB-ELBO), but we believe that this can be reduced to $\gO(TN)$ by applying the alias method \citep{walker1974new}.
We have not implemented the alias method, because the computational cost of decoding/encoding ancestral indices only accounted for a small fraction of the overall cost in our experiments.

The neural net computation and the encoding/decoding operations in BB-IS and BB-SMC can also be parallelized over particles, which in an ideal implementation would result in $\gO(1)$ and $\gO(T)$ time for the two methods, respectively. However, BB-AIS is not easily parallelizable due to its sequential nature. To test parallel performance, we have written an end-to-end parallelized version of BB-IS, using the JAX framework \citep{bradbury2018}, results are shown in Figure \ref{fig:time_plot}.

 \section{Experimental Details}
\label{appendix:experimental_details}

\subsection{Lossless Compression on Synthetic Data}
\label{appendix:lossless_toy}
\paragraph{Mixture Model}
We used a mixture model with 1-dimensional observation and latent variables. The alphabet sizes of the observation and the latent were 64 and 256 respectively. The data generating distribution (prior and conditional likelihood distributions) counts were i.i.d.\ sampled random integers from the range $[1, 20]$ and then normalized. All the coders got access to the true data generating distribution of the model and used a uniform approximate posterior distribution. All coders were evaluated using a message consisting of 5000 symbols i.i.d.\ sampled from the model to compute the total and net bitrates. The ideal bitrate of each coder was computed by its corresponding (empirical) negative variational bound by resampling the $N$ particle system 100 times and averaging (except for the ELBO bound, which can be computed exactly).

\paragraph{Hidden Markov Model}
We used a hidden Markov model (HMM) with 1-dimensional observation and latent variables. The alphabet sizes of the observation and the latent were 16 and 32 respectively. The number of timesteps was set to 10. The data generating distribution (prior/transition/emission distributions) counts were i.i.d. sampled random integers from the range $[1, 20]$ and then normalized. All the coders got access to the true data generating distribution of the model and used a uniform approximate posterior distribution. All coders were evaluated using a message consisting of 5000 sequences i.i.d.\ sampled from the model to compute the total and net bitrates. The ideal bitrate of each coder was computed by its corresponding (empirical) negative variational bound of the sampled message, as with the mixture model. The entropy of the model was also computed empirically by the negative marginal likelihood of the sampled message which was computed by the forward algorithm.

\begin{figure}[h]
    \centering
    \hspace{0.03\textwidth}
    \begin{subfigure}[b]{0.45\textwidth}
    \centering
    \includegraphics[width=\linewidth]{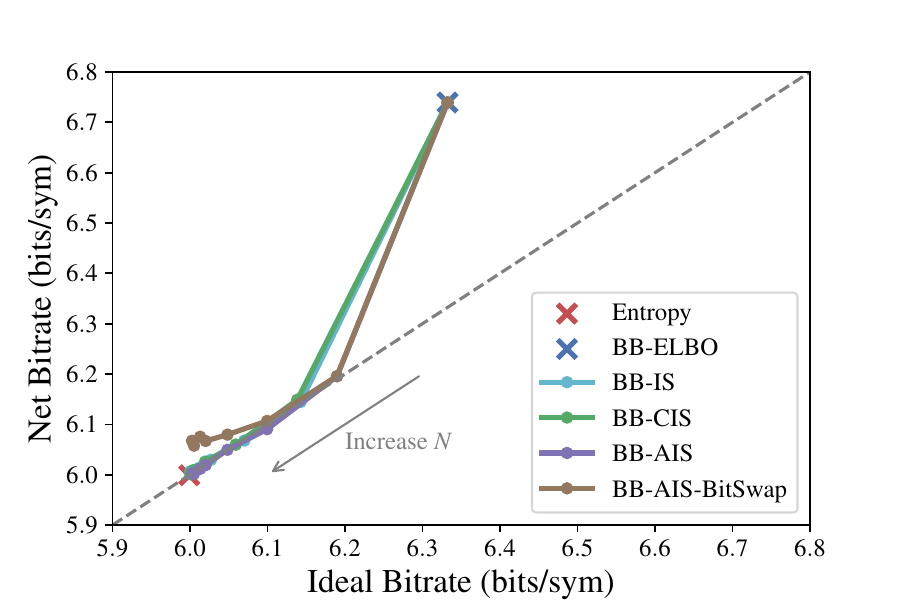}
    \caption{As $N \rightarrow \infty$, the net bitrates of all coders expect BB-AIS-BitSwap converge to the entropy on the toy mixture model. BB-AIS-BitSwap suffers from the dirty bits issue.}
    \label{fig:cleanliness_plot_toy_mixture_all}
    \end{subfigure}
    \hfill
    \begin{subfigure}[b]{0.45\textwidth}
    \centering
    \includegraphics[width=\linewidth]{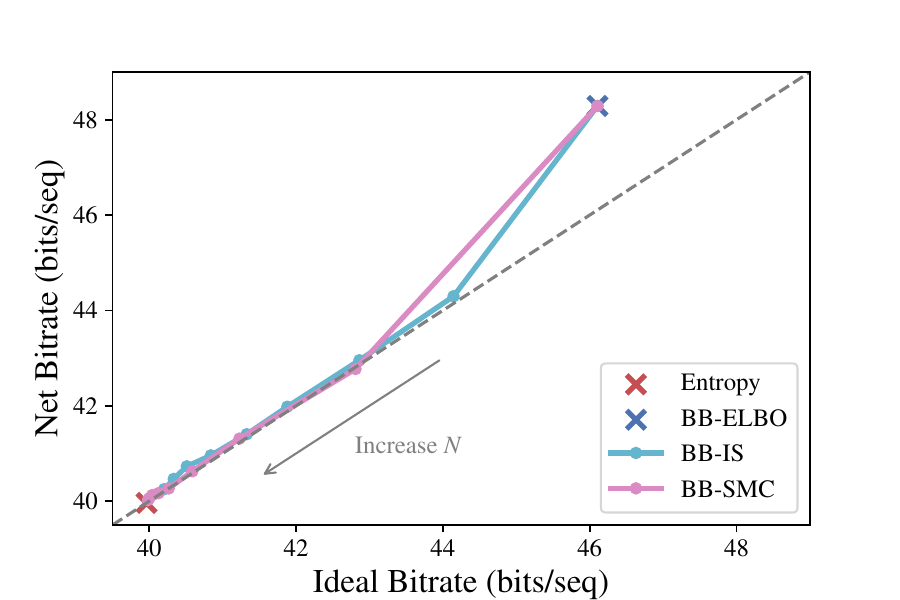}
    \caption{As $N \rightarrow \infty$, the net bitrates of BB-IS and BB-SMC coders converge to the entropy on the toy HMM. Both coders do not suffer from dirty bits issue.}
    \label{fig:cleanliness_plot_toy_hmm}
    \end{subfigure}
    \vspace{-1.0em}
    \caption{(a) \& (b): The cleanliness plots of all evaluated coders on the toy mixture model and the toy hidden Markov model. The experimental setups were the same as with Fig.  \ref{fig:cleanliness_plot_toy_mixture_only_ais}. and Fig, \ref{fig:convergence_plot_toy_hmm}, respectively. }
    \label{fig:cleanliness_plots_all}
    \hspace{0.03\textwidth}
\end{figure}

\paragraph{Additional Results} We include the cleanliness plots of all evaluated coders on the toy mixture and the toy HMM model in Fig. \ref{fig:cleanliness_plots_all}.

\subsection{Lossless Compression on Images}
\label{appendix:lossless_image}
\paragraph{Datasets}
We used two datasets for benchmarking lossless image compression: EMNIST \citep{cohen2017emnist} and CIFAR-10. EMNIST dataset extends the MNIST dataset to handwritten digits. There are 6 different splits provided in the dataset and we used two of them in our experiments: MNIST and Letters. The EMNIST-MNIST split mimics the original MNIST dataset which contains 60,000 training and 10,000 test examples. The EMNIST-Letters split contains 124800 training and 20800 test examples.
Both two splits were dynamically binarized following \citet{salakhutdinov2008quantitative}. Specifically, the observations were randomly sampled from the Bernoulli distribution with expectations equal to the real pixel values. For the CIFAR-10 dataset, no additional preprocessing was applied.

\paragraph{Model}
For the EMNIST datasets, we used the VAE model with 1 stochastic layer as in \citet{burda2015importance}. The VAE model had 50 latents with a standard Gaussian prior and factorized Gaussian approximate posterior. The conditional likelihood distribution was modeled by the Bernoulli distribution which fit with the binarized observations. The training procudure was the same as \citet{burda2015importance}. For the CIFAR-10 dataset, we used the VQ-VAE model \citep{oord2017neural} with discrete latent variables. Categorical distributions were used for both the latents and the observations. We followed the experimental setup in \citet{sonderby2017continuous} and used the VQ-VAE model with 8 latent vairables per spatial dimension. The VQ-VAE model was trained with Gumbel-Softmax \citep{jang2016categorical, maddison2016concrete} relaxation with a tempreture of 0.5 and a minibatch size of 32. The ADAM optimizer was used for training the model and the learning rate was tuned over $\{1 \times 10^{-4}, 5 \times 10^{-4}, 1 \times 10^{-3}\}$.

\paragraph{Discretization}
\label{appendix:lossless_image:discretization}
To perform bits-back coding with continuous latent variables, we need to discretize the latent space and approximate the continuous prior and the approximate posterior with their discretized variants using the same set of bins. In \citet{mackay2003information} and \citet{townsend2019practical}, they showed that the continuous latents could be discretized up to an arbitrary precision $\delta_z$ without affecting the \textit{net} bitrate as we could also get those extra bits due to discretization ``back". The notable effect is for the \textit{total} bitrate since the initial bit cost would scale with the $-\log \delta_z$, which means that we prefer to use a reasonably small precision in practice.
In our EMNIST experiments, continuous latent variables were used for the VAE model and need to be discretized for compression. We used the maximum entropy discretization in \citet{townsend2019practical} and discretized the latent space into bins that had equal \textit{mass} under the standard Gaussian prior for all the coders.

\paragraph{Baselines}
\label{appendix:lossless_image:baseline}
We included amortized-iterative inference method \citep{yang2020improving} which also improved the compression rate by bridging the gap of the ELBO bound and the marginal likelihood as a baseline in our experiments. When compressing a data example, it first initializes the \textit{local} variational parameters (e.g., the mean and variance for factorized Gaussian approximate posterior in our experiments) from the trained VAE inference model. Then it optimizes the ELBO objective over the local variational parameters with stochastic gradient decent and uses the optimized variational parameters for compression. This method introduces expensive computation at the compression stage. In our experiments, we kept the number of optimization steps equal to the number of particles $N$ to roughly match the computation budget in terms of the number of quries to the VAE model. However, even so the computation cost of this method is more expensive than our BB-IS/BB-CIS coders because 1) BB-IS/BB-CIS coders with $N$ particles introduce $N$ queries to the VAE generative model and $1$ query to the VAE inference model, while amortized-iterative inference with $N$ optimization steps requires $N$ quiries to the whole VAE model and $N$ back-propagation steps; 2) the computation of the BB-IS/BB-CIS coders can be potentially parallelized over the particles, but the computation of amortized-iterative inference needs to be done in $N$ sequential optimization steps. In our experiments, we used ADAM optimizer for optimizing the local variational parameters and the learning rate was tuned in the range $\{5\times 10^{-4}, 1\times 10^{-3}, 5\times 10^{-3}, 1\times 10^{-2}, 5\times 10^{-2}\}$.

\paragraph{Additional Results} We include some additional results for lossless compression on images here.

\label{appendix:lossless_image:additional}

\emph{Coupling does not hurt net bitrates\quad} To study the potential trade-off of applying our BB-CIS coder, we compared the compression performance of BB-IS and BB-CIS with 50 particles on EMNIST datasets in Table \ref{table:emnist_bb_is_with_bb_qmc}. Although the BB-CIS coder adopts a near-random sampling strategy, its ideal bitrate and net bitrate match those of the BB-IS coder. And it is effective for reducing the initial bit cost, as the gap between the total bitrate and the net bitrate is much smaller than that of the BB-IS coder. These observations illustrate that we can use BB-CIS as a drop-in replacement for BB-IS nearly for free.
\begin{table}[h]
\caption{BB-CIS is better than BB-IS in terms of total bitrate while matching the ideal and net bitrates. Comparison was done on the EMNIST-MNIST and EMNIST-Letters test sets. All bitrates were measured in bits/dim.}
\label{table:emnist_bb_is_with_bb_qmc}
\begin{center}
\begin{small}
\begin{tabular}{lccccccc}
\toprule
\multirow{2}{*}{Method}    & \multicolumn{3}{c}{MNIST} &    & \multicolumn{3}{c}{Letters} \\
\cmidrule{2-4}\cmidrule{6-8} &    Ideal   &   Net   &   Total  &    &  Ideal    &   Net   &   Total \\
\midrule
BB-IS (50)   &   0.227    &   0.228    &   0.230    &       &    0.238   &   0.239    &   0.241    \\
BB-CIS (50)   &  0.227     &    0.228   &  \textbf{0.228}     &       &   0.238    &   0.239     &  \textbf{0.239}     \\

\bottomrule
\end{tabular}
\end{small}
\end{center}
\vskip -0.1in
\end{table}

\emph{Combine BB-IS with amortized-iterative inference\quad} The compression performance of our BB-IS coder can be further improved by applying amortized-iterative inference at the compression stage. Similar to the original amortized-iterative inference, one can initialize the local variational parameters from the trained VAE inference model and optimize the IWAE objective with $N$ particles (as opposed to the ELBO objective) over them. In Table \ref{table:emnist_bb_is_with_iter}, we compared the \textit{net} bitrate of combining the BB-IS coder and the amortized-iterative inference method. We observed that the amortized-iterative inference method could further boost the compression rate of our BB-IS coder.

\begin{table}[h]
\caption{The compression performance of BB-IS can be improved by combining with amortized-iterative inference method. The net bitrates (bits/dim) are shown in the table.}
\label{table:emnist_bb_is_with_iter}
\begin{center}
\begin{small}
\begin{tabular}{lcccc}
\toprule
        &   \multicolumn{2}{c}{MNIST}    &   \multicolumn{2}{c}{Letters}  \\
        \cmidrule(lr){2-3} \cmidrule(lr){4-5} & w/ IF & w/o IF & w/ IF & w/o IF \\
\midrule
    BB-IS (1)   &   0.236    &   0.233   &    0.250   &    0.246 \\
    BB-IS (5)   &   0.231   &   0.229   &    0.243   &       0.241  \\
    BB-IS (50)  &   0.228   &   \textbf{0.226}   &   0.239   &   \textbf{0.237}   \\
\bottomrule
\end{tabular}
\end{small}
\end{center}
\vskip -0.1in
\end{table}

\emph{Comparison with amortized-iterative inference in OOD settings\quad} We also compare the out-of-distribution (OOD) compression performance of BB-IS with amortized-iterative inference, as shown in Table \ref{table:emnist_transfer_bb_is_iter}. We observed that with roughly the same computation budget, BB-IS outperforms the amortized-iterative inference method in OOD compression settings. The most improved compression rates can be achieved when BB-IS is combined with it (denoted as BB-IS (50)-IF (50)).

\begin{table}[h]
\caption{BB-IS outperforms amortized-iterative inference in OOD compression setting. The net bitrates (bits/dim) are shown in the table.}
\label{table:emnist_transfer_bb_is_iter}
\begin{center}
\begin{small}
\begin{tabular}{lcccc}
\toprule
\multicolumn{1}{c}{Trained on} & \multicolumn{2}{c}{MNIST} & \multicolumn{2}{c}{Letters} \\ \cmidrule(lr){1-1} \cmidrule(lr){2-3} \cmidrule(lr){4-5}
\multicolumn{1}{c}{Compressing} & MNIST & Letters  & MNIST & Letters\\
\midrule
BB-ELBO & 0.236 & 0.310  & 0.257 & 0.250 \\
BB-ELBO-IF (50) & 0.233 & 0.294 & 0.252 & 0.246 \\
BB-IS (50) & 0.228 & 0.280 & 0.244 & 0.239 \\
BB-IS (50)-IF (50) & 0.227 & 0.272 & 0.241 & 0.237 \\
\bottomrule
\end{tabular}
\end{small}
\end{center}
\vskip -0.1in
\end{table}

\emph{Comparison with other existing methods\quad} We include the comparison of BB-IS with existing neural lossless compression baselines on CIFAR-10, as shown in Table \ref{table:cifar_other_baselines_compare}. Note that our McBits coders are not directly comparable with these methods since all of these assume different computational regimes or exploit
\emph{distinct} model classes from ours (e.g., VQ-VAE), but we still include the comparison to provide context for readers.

\begin{table}[h]
\caption{The comparison of total bitrates of BB-IS with other neural compression baselines on CIFAR-10. We emphasize that these methods are not directly comparable since the model classes used are very different, and the comparison between BB-IS and BB-ELBO within the same model class (i.e., VQ-VAE) is meaningful.}
\label{table:cifar_other_baselines_compare}
\begin{center}
\begin{small}
\begin{tabular}{cc}
\toprule
    Method  &   CIFAR-10 \\
\midrule
    gzip    &   7.37    \\
    bzip2   &   6.98    \\
    lzma    &   6.09    \\
    PNG     &   5.89    \\
    JPEG    &   5.20    \\
    WebP    &   4.61    \\
    FLIF    &   4.37    \\
\midrule
    REC \citep{flamich2020rec}    &   4.18    \\
    BitSwap \citep{kingma2019bit} &   3.82    \\
    IDF \cite{hoogeboom2019integer}     &   \textbf{3.34}    \\
\midrule
    BB-ELBO           &   4.90 \\
     BB-IS (10)        &   4.81 \\
\bottomrule
\end{tabular}
\end{small}
\end{center}
\vspace{-1.0\baselineskip}
\end{table}

\subsection{Lossless Compression on Sequential Data}

\paragraph{Datasets}

We used 4 polyphonic music datasets to evaluate the compression performance of the BB-SMC coder on sequential datasets: Nottingham folk tunes, the JSB chorales, the MuseData library of classical piano and orchestral music, and the Piano-midi.de MIDI archive \citep{boulanger2012modeling}.
All datasets  were composed of sequences of binary 88-dimensional vectors representing active  musical notes at one timestep.
For all datasets, we imitated the experimental setup presented in \citet{maddison2017filtering}.
 We used the same train/validation/test split and echoed their data preprocessing.
 The only difference was that we used a chuncked version of these datasets where each sequence was chunked to sub-sequences with maximum length of 100. We found it slightly improved the model performance and significantly reduced the initial bit cost (as the original maximum sequence length was very large).

\paragraph{Models}
All models were based on the variational RNN architecture \citep{chung2015recurrent}. All distributions over latent variables were factorized Gaussians, and the output distributions were factorized Bernoullis for binary observations on 4 polyphonic music datasets. JSB models were trained with 32 hidden units, Muse-data with 256, and all other models with 64 units. For each aforementioned dataset, there was one model trained with the ELBO, IWAE, and FIVO objectives, respectively.
All models were trained with 4 particles, a batch size of 4 and the Adam optimizer with learning rate $3 \times 10^{-5}$.
All models were initially evaluated on the validation set, which allowed for early stopping.
In Table \ref{table:model_evaluation_piano},
we present our models' performance in nats and bits to allow for easy comparison of generative modelling and compression literature.

\begin{table}[ht]
\centering
\caption{Sequential model evaluation: we trained VRNN models on the Nottingham, JSB, Musedata and piano-midi.de datasets. For each dataset, we trained 3 VRNN models with the ELBO, IWAE and FIVO objectives, respectively. All models were trained with 4 particles.
Our models were trained in an identical fashion as with Table 5 in \citet{maddison2017filtering}. For comparison, we include the estimated data log-likelihood as model evaluation metric. We estimated the log-likelihood by computing the maximum of the ELBO, IWAE and FIVO bound with 128 particles. We include this metric for better comparison to other work. However for this work, this bound is not relevant. Relevant metrics include the respective bounds in nats or bits per time step. }
\label{table:model_evaluation_piano}

\vskip 0.15in
\begin{small}

\begin{tabular}{llcrrrrrrrrrrrrrr}
\toprule
Training              & Evaluation  & \multirow{2}{*}{Unit} & \multicolumn{2}{c}{Notingham}                        & \multicolumn{1}{c}{} & \multicolumn{2}{c}{JSB}                              & \multicolumn{1}{c}{} & \multicolumn{2}{c}{Musedata}                         & \multicolumn{1}{c}{} & \multicolumn{2}{c}{Piano-midi.de}                    \\ \cmidrule{4-5} \cmidrule{7-8} \cmidrule{10-11} \cmidrule{13-14}
Objective             & Metric      &                       & \multicolumn{1}{c}{Train} & \multicolumn{1}{c}{Test} & \multicolumn{1}{c}{} & \multicolumn{1}{c}{Train} & \multicolumn{1}{c}{Test} & \multicolumn{1}{c}{} & \multicolumn{1}{c}{Train} & \multicolumn{1}{c}{Test} & \multicolumn{1}{c}{} & \multicolumn{1}{c}{Train} & \multicolumn{1}{c}{Test} \\
\midrule
\multirow{3}{*}{ELBO} &  $-\log p(x)$ & nats/step     &  3.49            & 4.06   & &8.05                     &8.67                    & &6.50                    &7.33                   & &   7.30        &  7.92                 \\
                      & $-$ELBO        & nats/step     &  3.50            & 4.07   & &8.07                     &8.67                    & &6.53                    &7.38                   & &   7.31        &  7.93                 \\
                      & $-$ELBO        & bits/step     &  5.05            & 5.87   & &11.64                    &12.51                   & &9.41                    &10.65                  & &  10.55        &11.44                  \\ \hline
\multirow{3}{*}{IWAE} &  $-\log p(x)$ & nats/step     &  2.51            & 3.03   & &7.50                     &8.13                    & &6.40                    &7.33                   & &  7.25         & 7.88                       \\
                      & $-$IWAE        & nats/step     &  2.63            & 3.24   & &7.65                     &8.36                    & &6.42                    &7.38                   & &  7.89         & 7.89                        \\
                      & $-$IWAE        & bits/step     &  3.79            & 4.67   & &11.04                    &12.06                   & &9.26                    &10.65                  & & 11.38         & 11.38                  \\ \hline
\multirow{3}{*}{FIVO} &  $-\log p(x)$ & nats/step     & 2.50             & 3.02   & &6.41                     &7.24                    & &5.82                    &6.46                   & &  6.94         & 7.70                      \\
                      & $-$FIVO        & nats/step     & 2.60             & 3.20   & &6.59                     &7.50                    & &5.97                    &6.64                   & &  6.99         & 7.76                       \\
                      & $-$FIVO        & bits/step     & 3.75             &4.62    & &9.51                     &10.82                   & &8.61                    &9.58                   & &  10.08        & 11.20                    \\
\bottomrule
\end{tabular}

\end{small}
\end{table}

\paragraph{Additional Results}
\label{appendix:lossless_sequential:additional}
We include some additional results for lossless compression on sequential data here.

We compared our coders with benchmark lossless compression schemes in Table \ref{table:seqential_data_baselines_compare}. Our coders were comparable with those baselines and the BB-SMC coder outperformed all the baselines on the JSB dataset. Note that the compression performance of our coders is bottlenecked by the simple VRNN architecture that we used in our experiments. And we suppose that with more powerful and better trained VRNN models, our coders could outperform those benchmark schemes. However, our main focus is to compare the compression performance of our BB-SMC coder and the BB-ELBO/BB-IS coders with the same VRNN architectures to show the effectiveness of BB-SMC for compressing sequential data, and this is clearly illustrated by the results.
\begin{table*}[!ht]
\caption{The comparison of net bitrate (bits/timestep) with benchmark lossless compression schemes on sequential data benchmarks.}
\label{table:seqential_data_baselines_compare}
\begin{center}
\begin{small}
\begin{tabular}{ccccc}
\toprule
    Method         &   Musedata    &   Nottingham   &   JSB     &   Piano-midi.de     \\
\midrule
    gzip     &   11.01    &   3.86    &   13.94    &   9.46   \\
    bz2     &    11.25      &   \textbf{2.95}    &    {11.97}   &   10.67    \\
    lzma     &   \textbf{8.44}    &   3.12    &    12.78   &  \textbf{7.27}     \\
\midrule
    BB-ELBO     &    10.66   &    5.87   &   12.53    &    11.43   \\
    BB-IS (4)     &     10.66      &  4.86     &   12.03    &   11.38 \\
    BB-SMC (4)     &   {9.58}    &   {4.76}    &   \textbf{10.92}    &   {11.20}   \\
\bottomrule
\end{tabular}
\end{small}
\end{center}
\vskip -0.1in
\end{table*}
\paragraph{Discussion of Initial Bit Cost}
\label{appendix:lossless_sequential:discussion}
The initial bit cost of compressing sequential data using (Monte Carlo) bits-back algorithms scales linearly with both the sequence length and the number of particles, i.e., $\gO (NT)$.
The original four polyphonic music datasets have a special characteristic that the average and maximum sequence lengths are large but the number of sequences is small, which means that the initial bit cost is huge but cannot be sufficiently amortized. Thus, if we compress the original datasets without chunking sequences, the \textit{total} bitrate will be much larger than the \textit{net} bitrate.
For example, there are only 124 sequences in the Musedata test set but the average length is 519 and the maximum length is 4273. As a result, the total bitrates of the BB-ELBO coder and the BB-SMC coders for compressing the original dataset are 136.81 and 544.40 bits/timestep respectively and much larger than their net ones.
Therefore, we chose to chunk long sequences to short ones of a predefined maximum length (100) and compress them independently, which could effectively decrease and amortize the initial bit cost. When compressing the chunked dataset, the total bitrates of BB-ELBO and BB-SMC reduce to 12.83 and 21.39 bits/timestep respectively.
As for the initial bit cost caused by the particles, we can also use the coupled variant of BB-SMC (aka BB-CSMC) for compressing sequential data.

\subsection{Lossy Compression on Images}
\label{appendix:lossy_image}

\paragraph{Lossy Compression Setup} We used the binarized EMNIST datasets to benchmark the lossy compression performance. We considered the lossy compression setup with hierarchical VAE models \citep{balle2018variational, minnen2018joint}. Specifically, the compressing data $x$ is transformed by trained hierarchical inference models $f_l$ and $f_h$ with parameters $\phi_l$ and $\phi_h$ to produce discretized latent $y$ and hyperlatent $z$ as $y=\lfloor \mu_y^f \rceil$ and $z=\lfloor \mu_z^f \rceil$, where $\mu_y^f=f_l(x;\phi_l)$ and $\mu_z^f=f_l(y;\phi_h)$ are their continuous representations. In our experiments, the latent is rounded to the nearest integer and the hyperlatent is discretized by the maximum entropy discretization scheme introduced in \ref{appendix:lossless_image:discretization} for lossless compression. Then the latent $y$ and the hyperlatent $z$ are compressed with bits-back coding as in lossless compression. On the decoder side, both $y$ and $z$ can be losslessly recovered and the reconstructed data $\hat{x}$ is transformed from $y$ using the generative models $g_l$ and $g_h$ with parameters $\theta_l$ and $\theta_h$.

\paragraph{Model} We used the VAE model with 2 stochastic layers in \citet{burda2015importance} with several modifications based on \citet{balle2018variational} for adapting to the lossy compression setup. Specifically, the approximate posterior distribution of the latent was a uniform distribution centered at $\mu_y$, i.e.,  $q(y|x)=\uniform(\mu_y^f-\frac{1}{2}, \mu_y^f+\frac{1}{2})$, which is a differentiable substitute for rounding during training. The conditional likelihood distribution of the latent should also support quantization which was a factorized Gaussian distribution convolved with a standard uniform $p(y|z) = \gaussian(\mu_y^g, {\sigma_y^g}^2) * \uniform(-\frac{1}{2}, \frac{1}{2})$, where $(\mu_y^g, \sigma_y^g)=g_h(z;\theta_h)$. The convolved distribution agrees with the discretized distribution on all integers \citep{balle2016end, balle2018variational}. This is important because the discretization of the latent would affect the compression rate which should be taken into account during training. In contrast, the disretization of the hyperlatent does not affect the compression rate as a result of getting bits back (see the discussion in \ref{appendix:lossless_image:discretization}), we kept the distribution over the hyperlatent unchanged as in  \citet{burda2015importance}. Specifically, its approximate posterior distribution $q(z|y)$ was a factorized Gaussian distribution centered at $\mu_z^f$ and its prior distribution $p(z)$ was a standard Gaussian. The conditional likelihood distribution was kept as a factorized Bernoulli for binary observations.

\paragraph{Training} The loss function of training the hierarchical VAE model is a relaxed rate-distortion objective:
\begin{equation}
    \mathcal{L_\lambda}(\theta_l, \theta_h, \phi_l, \phi_h)=\expect_{q(y|x)q(z|y)}\left[\underbrace{-\lambda \log(x|y)}_{\text{weighted distortion}} \underbrace{- \log \frac{p(y,z)}{q(z|y)}}_{\text{rate as ELBO}} + \underbrace{\cancel{\log q(y|x)}}_{0} \right]
\end{equation}
The distortion is measured as the negative log likelihood of the Bernoulli observations and the rate is measure as the negative ELBO marginalized over the hyperlatent. $\lambda$ is the hyperparameter that controls the rate-distortion trade-off. The last term is measured as $0$ since $q(y|x)$ is a uniform distribution. The above rate-distortion objective is very similar to the objective function in $\beta$-VAE \citep{higgins2016beta} and can be optimized with the reparameterization trick. Note that the ELBO rate term can be changed to the IWAE objective with multiple particles.

In our experiments, we trained the models on the binarized EMNIST-MNIST dataset and evaluated on both EMNIST-MNIST and EMNIST-Letters test sets (for evaluating the performance in the transfer setting). we trained the model with the above loss function using the IWAE objective with $M \in \{1, 5, 50\}$ particles as the rate term. For each setup, we trained models with different $\lambda$ values in the range $\{1.0, 1.5, 2.0, 2.5, 3.0, 4.0, 5.0, 6.0, 7.0, 7.5, 8.0, 9.0, 10.0, 12.5, 15.0, 17.5, 20.0\}$. For training each model, we tuned the learning rate in the range $\{5\times10^{-3}, 2.5\times10^{-3}, 1\times10^{-3}, 7.5\times10^{-4}, 5\times10^{-4}\}$.

\paragraph{Additional Results}
\label{appendix:lossy_image:additional}
We include some additional results for lossy image compression here.

\begin{figure}[h]
    \centering
    \begin{subfigure}[b]{0.45\textwidth}
        \includegraphics[width=\linewidth]{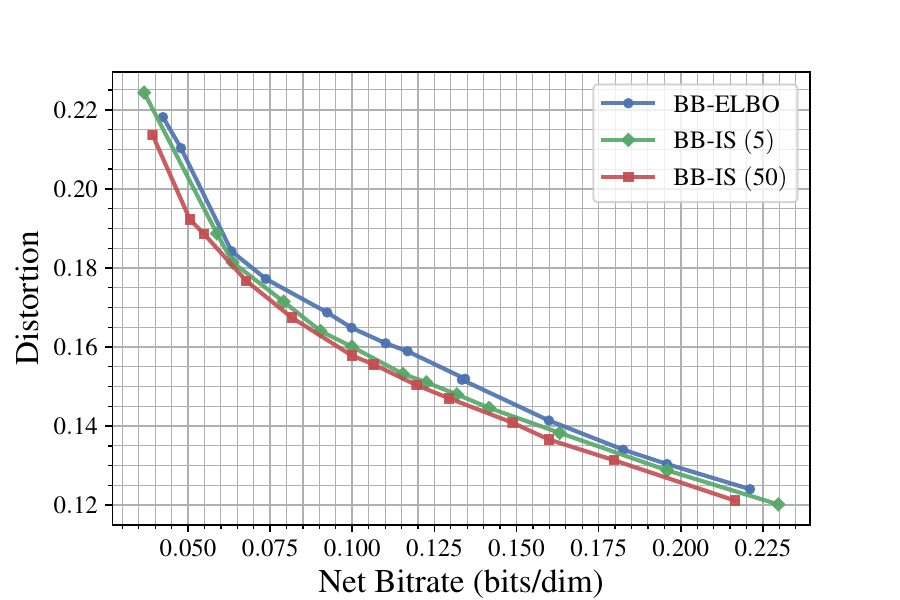}
        \caption{}
        \label{fig:lossy_mnist_bb_is_rd_curve}
    \end{subfigure}
    \begin{subfigure}[b]{0.45\textwidth}
        \includegraphics[width=\linewidth]{figures/lossy_mnist_bb_is_rate_saving_curve.pdf}
        \caption{}
        \label{fig:lossy_mnist_bb_is_rate_saving_curve_2}
    \end{subfigure}
    \begin{subfigure}[b]{0.45\textwidth}
    \centering
    \includegraphics[width=\linewidth]{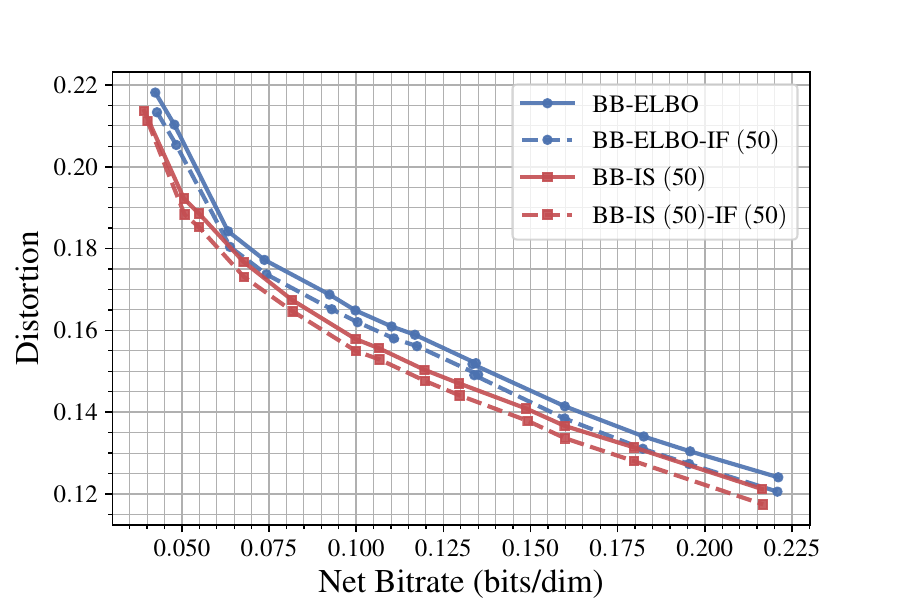}
    \caption{}
    \label{fig:lossy_mnist_with_iter_rd_curve}
    \end{subfigure}
    \begin{subfigure}[b]{0.45\textwidth}
    \centering
    \includegraphics[width=\linewidth]{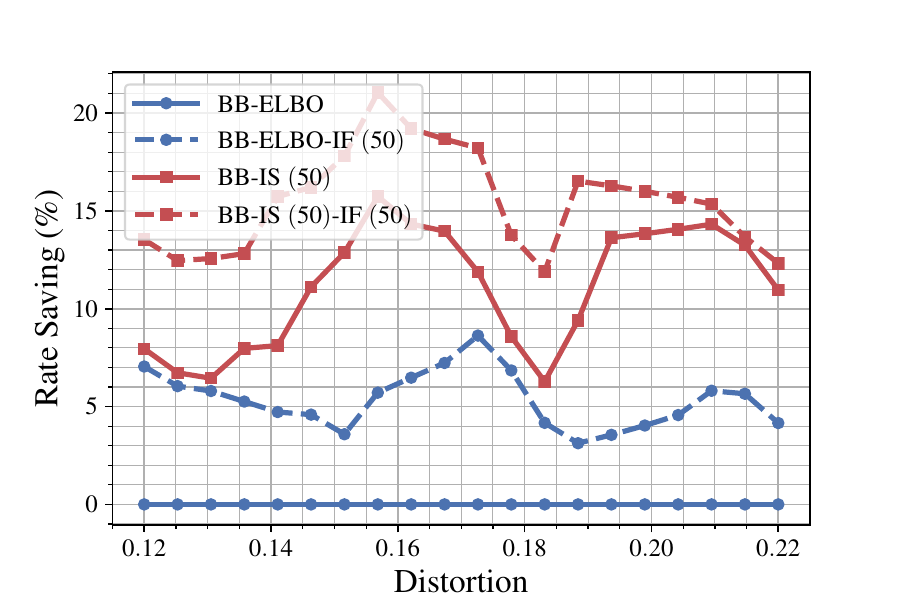}
    \caption{}
    \label{fig:lossy_mnist_with_iter_rate_saving_curve}
    \end{subfigure}
    \vspace{-1.0em}
    \caption{The lossy compression performance on EMNIST-MNIST test set with models trained on EMNIST-MNIST training set. (a) \& (b): The rate-distortion curve and the rating saving curve for comparing BB-IS and BB-ELBO. Our BB-IS coder achieves better rate-distortion trade-off than BB-ELBO. (c) \& (d): The rate-distortion curve and the rating saving curve with amortized-iterative inference.  With fixed computation budget, BB-IS outperforms amortized-iterative inference and can be combined with it to further improve the performance. We measure the rate savings ($\%$) relative to BB-ELBO for fixed distortion values. }
    \label{fig:lossy_mnist_all}

\end{figure}

We include the rate-distortion curve and the rate saving curve evaluated on the EMNIST-MNIST test set in Fig. \ref{fig:lossy_mnist_bb_is_rd_curve} and Fig. \ref{fig:lossy_mnist_bb_is_rate_saving_curve_2}, respectively. We found that BB-IS achieved better rate-distortion trade-off than BB-IS and achieved more than 15 \% rate savings in some setups.

We also evaluated the performance of applying amortized -iterative inference \citep{yang2020improving} in lossy compression in Fig. \ref{fig:lossy_mnist_with_iter_rd_curve} \& \ref{fig:lossy_mnist_with_iter_rate_saving_curve}. The main purposes were to: 1) compare the amortized-iterative inference and our BB-IS coder with similar computation budget; 2) illustrate the potential of combining amortized-iterative inference with our BB-IS coder. Specifically, we used 50 optimization steps for amortized-iterative inference to roughly match the computation budget (see discussion in \ref{appendix:lossless_image:baseline}) with our BB-IS coder with 50 particles (denoted as BB-IS (50)). We used a 2-stage amortized-iterative inference similar to \citet{yang2020improving} and each stage contained 25 optimization steps. In the first stage, both the local variational parameters of the latent and the hyperlatent were optimized with the rate-distortion objective. In the second stage, the local variational parameters of the latent were fixed while those of the hyperlatent were optimized with the rate term (i.e., negative ELBO) of the rate-distortion objective. The learning rate was tuned in the range $\{1 \times 10^{-2}, 5 \times 10^{-3}, 1 \times 10^{-3}\}$. This method is denoted as BB-ELBO-IF (50). We observed that BB-ELBO-IF (50) improved over BB-ELBO but underperformed our BB-IS (50) in terms of rate-distortion trade-off. We also combined BB-IS (50) with 50 optimization steps of amortized-iterative inference using negative IWAE as the rate term (denoted as BB-IS (50)-IF (50)), which we found to further improve the performance and achieved more than $20\%$ rate savings in some setups.

\begin{figure}[h]
    \centering
    \begin{subfigure}[b]{0.45\textwidth}
    \centering
    \includegraphics[width=\linewidth]{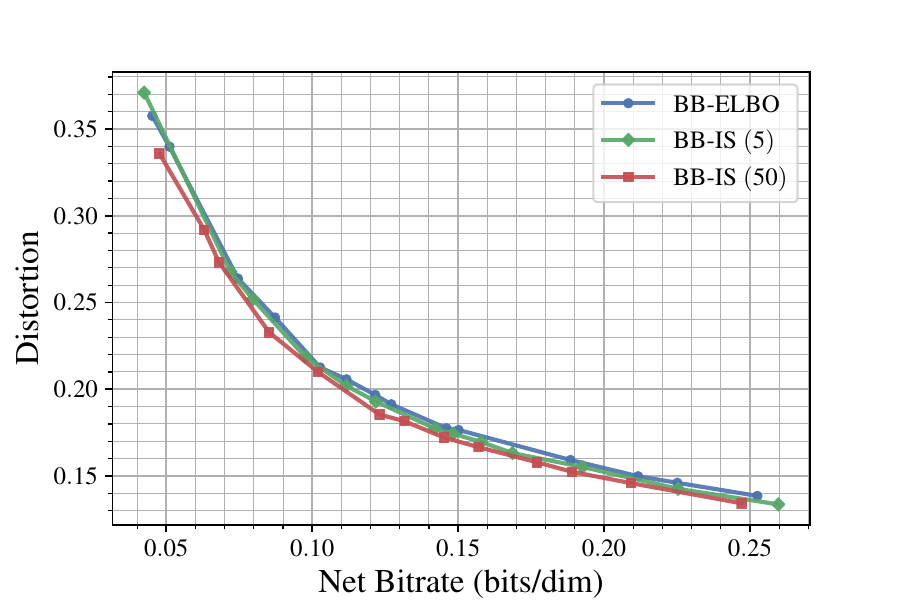}
    \caption{}
    \label{fig:lossy_letters_bb_is_rd_curve}
    \end{subfigure}
    \begin{subfigure}[b]{0.45\textwidth}
    \centering
    \includegraphics[width=\linewidth]{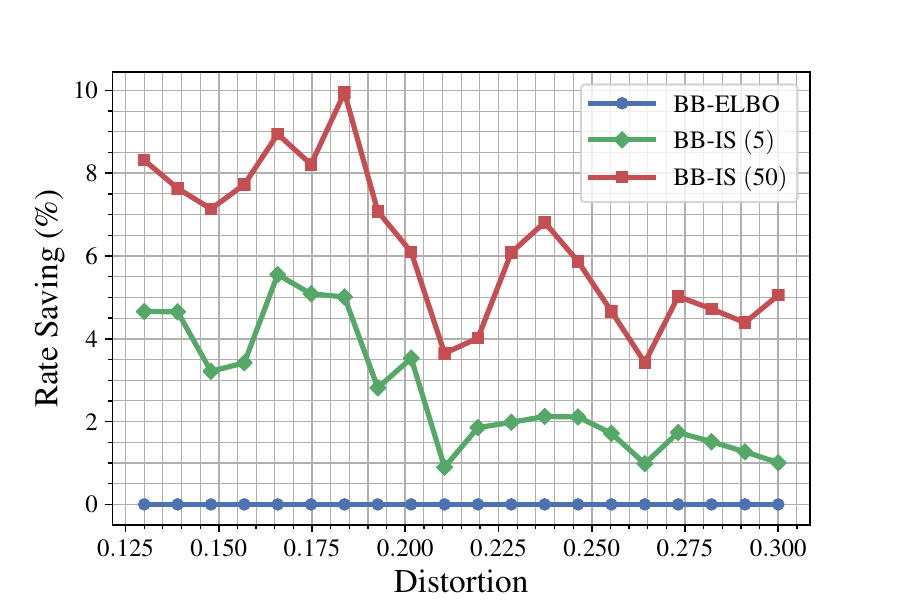}
    \caption{}
    \label{fig:lossy_letters_bb_is_rate_saving_curve}
    \end{subfigure} \\
    \begin{subfigure}[b]{0.45\textwidth}
    \centering
    \includegraphics[width=\linewidth]{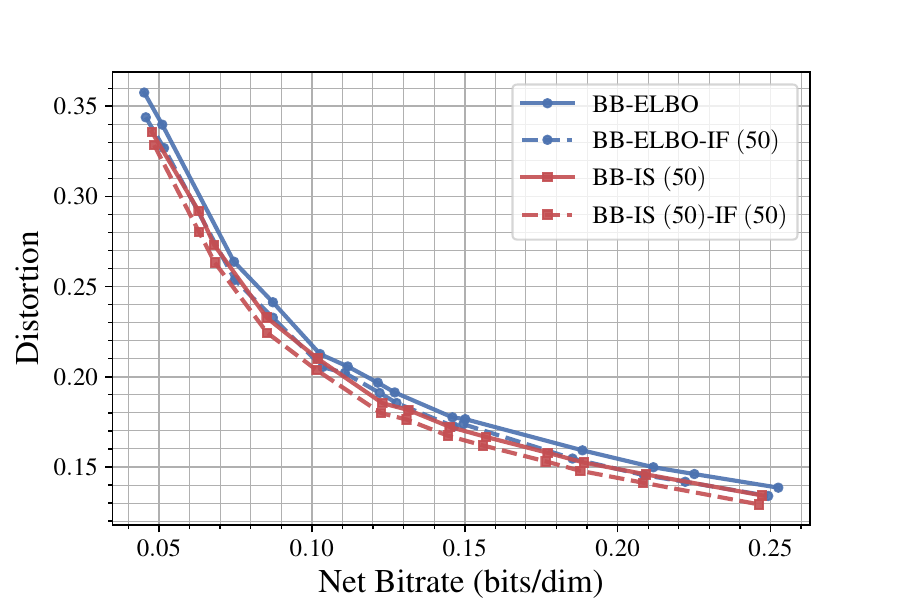}
    \caption{}
    \label{fig:lossy_letters_with_iter_rd_curve}
    \end{subfigure}
    \begin{subfigure}[b]{0.45\textwidth}
    \centering
    \includegraphics[width=\linewidth]{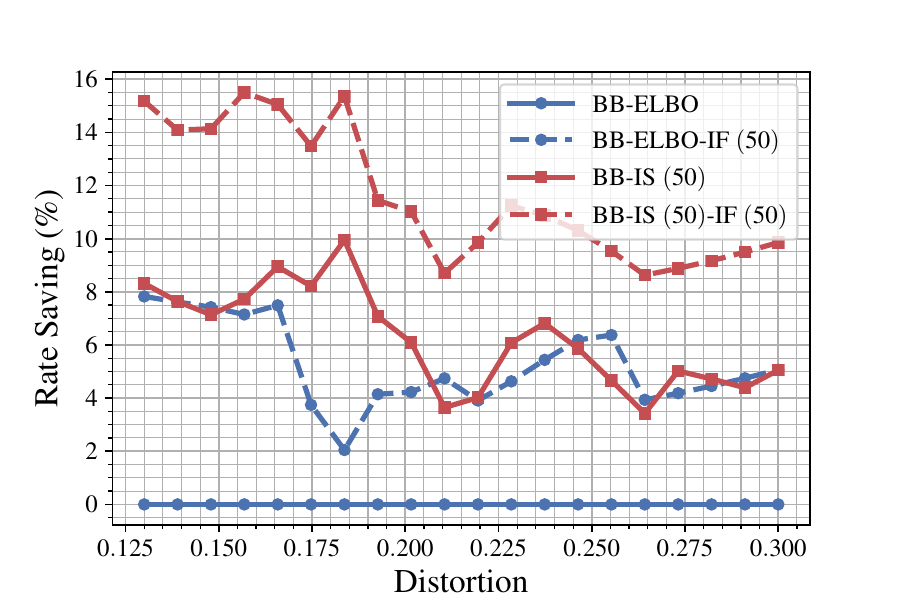}
    \caption{}
    \label{fig:lossy_letters_with_iter_rate_saving_curve}
    \end{subfigure}
    \vspace{-1.0em}
    \caption{The lossy compression performance on EMNIST-Letters test set with models trained on EMNIST-MNIST training set. We observe similar results as Fig. \ref{fig:lossy_mnist_all}, but the improvement of BB-IS is not as significant as Fig. \ref{fig:lossy_mnist_all} in this transfer setting. (a) $\&$ (b): The rate-distortion curve and the rating saving curve for comparing BB-IS and BB-ELBO. (c) $\&$ (d): The rate-distortion curve and the rating saving curve with amortized-iterative inference. }
    \label{fig:lossy_letters_all}

\end{figure}

We also evaluated the lossy compression performance in a transfer setting where we used models trained on the EMNIST-MNIST to compress EMNIST-Letters test set, as in Fig. \ref{fig:lossy_letters_all}. We observed that the improvement of BB-IS was not as significant as on the EMNIST-MNIST test set. This might be due to that although BB-IS could improve the rate term over BB-ELBO more on the EMNIST-Letters test set (as shown in Table \ref{table:vae_transfer} for lossless compression), the distortion term of the model trained with IWAE might not generalize well on the dataset of a slightly different distribution.

\end{document}